\documentclass[smallextended]{svjour3}       

\smartqed  
\usepackage[american]{babel}
\journalname{myjournal}
\usepackage{natbib}

\usepackage{amsmath,amssymb,array}
\usepackage{dsfont}
\newtheorem{proof-sketch}{Proof Sketch}

\usepackage{tikz}
\usetikzlibrary{arrows,chains,matrix,positioning,scopes}

\usepackage{xcolor}
\usepackage{graphicx}
\usepackage{xargs}
\usepackage{tikz}
\usetikzlibrary{arrows}
\usepackage{bm}
\usepackage[utf8]{inputenc}
\usepackage[longtable]{multirow}
\usepackage{booktabs}
\usepackage{colortbl}
\usepackage{csquotes} 
\usepackage{mathtools}
\usepackage{float} 
\usepackage{adjustbox}

\newcommand{\Xspace}{\mathcal{X}}                                           
\newcommand{\Yspace}{\mathcal{Y}}                                           
\newcommand{\D}{\mathcal{D}}                                                
\renewcommand{\xi}[1][i]{\mathbf{x}^{(#1)}}                                          
\newcommand{\yi}[1][i]{y^{(#1)}}                                            

\newcommand{\Hspace}{\mathcal{H}}														
\newcommand{\fh}{\hat{f}}                                                   
\newcommand{\risk}{\mathcal{R}}                                             
\newcommand{\riske}{\mathcal{R}_{\text{emp}}}                               

\ifdefined\N                                                                
\renewcommand{\N}{\mathds{N}}                                                
\else
  \newcommand{\N}{\mathds{N}}
\fi
\newcommand{\R}{\mathds{R}}                                                 
\ifdefined\C
  \renewcommand{\C}{\mathds{C}}                                             
\else
  \newcommand{\C}{\mathds{C}}
\fi

\newcommand{\sumin}{\sum\limits_{i=1}^n}											
\newcommand{\meanin}{\frac{1}{n} \sum\limits_{i=1}^n}			    				

\renewcommand{\P}{\mathds{P}}                                               
\newcommand{\E}{\mathds{E}}                                                 
\newcommand{\var}{\mathbb{V}}                                             

\renewcommand{\xi}{x^{(i)}}        

\newcommand{\F}{F}

\newcommand{\EF}{\E_{\F}}
\newcommand{\VF}{\var_{\F}}

\newcommand{\Exyperm}{\E_{\tilde{X_S}X_C Y}}
\newcommand{\Experm}{\E_{\tilde{X_S}X_C}}
\newcommand{\Exynorm}{\E_{X_S X_C Y}}
\newcommand{\Exnorm}{\E_{X_S X_C}}
\newcommand{\Var}{\mathbb{V}}

\newcommand{\EXY}{\E_{XY}}
\newcommand{\EXZ}{\E_{\tilde{X}_SX}}
\newcommand{\EXYZ}{\E_{\tilde{X}_SXY}}
\newcommand{\EXYF}{\E_{\F XY}}
\newcommand{\EXYZF}{\E_{\F\tilde{X}_SXY}}
\newcommand{\tpdp}{\E_{X_C}[f]}
\newcommand{\mpdp}{\E_{X_C}[\fh]}

\newcommand{\indep}{\rotatebox[origin=c]{90}{$\models$}}
\newcommand{\Li}{L^{(i)}}
\newcommand{\Lt}{\tilde{L}}
\newcommand{\Ltmean}{\bar{\Lt}^{(i)}}

\newcommand{\talpha}{t_{1-\tfrac{\alpha}{2}}}
\newcommand{\elPFI}{\overline{\widehat{PFI}}}
\newcommand{\elPD}{\overline{\widehat{PD}}}

\renewcommand{\riske}{\hat{\risk}}

\usepackage{color}
\usepackage{soul}

\begin{document}

\title{Relating the Partial Dependence Plot and Permutation Feature Importance to the Data Generating Process\thanks{This project is funded by the Bavarian State Ministry of Science and the Arts and coordinated  by  the  Bavarian  Research  Institute  for  Digital  Transformation  (bidt) and supported by the German Federal Ministry of Education and Research (BMBF) under Grant No. 01IS18036A, by the German Research Foundation (DFG) -- Emmy Noether Grant 437611051 to MNW, and by the Graduate School of Systemic Neurosciences (GSN) Munich. The authors of this work take full responsibilities for its content.}}

 \titlerunning{Relating PDP and PFI to the Data Generating Process}

\institute{F. Author \at
              first address \\
              Tel.: +123-45-678910\\
              Fax: +123-45-678910\\
              \email{fauthor@example.com}           
           \and
           S. Author \at
              second address
}

\author{%
Christoph Molnar
\and Timo Freiesleben
\and Gunnar König
\and Giuseppe Casalicchio
\and Marvin N. Wright
\and Bernd Bischl
}

%
%
\institute{
C. Molnar, T. Freiesleben, G. König, G. Casalicchio, B. Bischl \at
Ludwig-Maximilian University Munich,  Germany \and
  G. König \at
  University of Vienna, Austria \and
  C. Molnar, M. Wright \at
  Leibniz Institute for Prevention Research and Epidemiology -- BIPS, Bremen, Germany \and
  M. Wright \at
  University of Bremen, Germany
}

\date{Received: date / Accepted: date}

\maketitle              

\begin{abstract}
Scientists and practitioners increasingly rely on machine learning to model data and draw conclusions.
Compared to statistical modeling approaches, machine learning makes fewer explicit assumptions about data structures, such as linearity.
However, their model parameters usually cannot be easily related to the data generating process.
To learn about the modeled relationships, partial dependence (PD) plots and permutation feature importance (PFI) are often used as interpretation methods.
However, PD and PFI lack a theory that relates them to the data generating process.
We formalize PD and PFI as statistical estimators of ground truth estimands rooted in the data generating process.
We show that PD and PFI estimates deviate from this ground truth due to statistical biases, model variance and Monte Carlo approximation errors.
To account for model variance in PD and PFI estimation, we propose the learner-PD and the learner-PFI based on model refits, and propose corrected variance and confidence interval estimators.
\end{abstract}

\section{Introduction}
\label{sec:introduction}

Statistical models such as linear or logistic regression models are frequently used to learn about relationships in data.
Assuming that a statistical model reflects the data generating process (DGP) well, we may interpret the model coefficients in place of the DGP and draw conclusions about the data.
An important part of interpreting the coefficients is the quantification of their uncertainty via standard errors, which allows to separate random noise (non-significant coefficients) from real effects.
Statistical biases and violation of assumptions are well studied for many model classes, such as heterogeneous residuals, deviations from normality, and non-additivity for linear models \citep{fahrmeir2007regression}.

Increasingly, machine learning approaches such as gradient-boosted trees, random forests or neural networks are used instead of or in addition to statistical models.
Compared to statistical models that are driven by considerations of the data generating process, the machine learning approaches often lack a mapping between model parameters and properties of the DGP.
Due to the ability of many machine learning models to address highly non-linear relationships and interactions, they often outperform more restrictive statistical models.
Scientific applications of machine learning are widespread and range from modeling volunteer labor supply \citep{bair2013multivariable}, mapping fish biomass \citep{esselman2015landscape}, analyzing urban reservoirs \citep{obringer2018predicting}, identifying disease-associated genetic variants  \citep{boulesteix2020statistical}, and inferring behavior from smartphone use \citep{Stachl17680}.
In these scientific applications, the model is only the means to an end: a better understanding of the data generating process, in particular the conditional expectation of the target variables as a function of the features.

Model-agnostic interpretation methods \citep{Ribeiro2016b} are a (partial) remedy to the lack of interpretable parameters of more complex models.
Model-agnostic methods follow a general procedure of 1) sampling data, 2) manipulating this data, 3) predicting and 4) aggregating the predictions \citep{scholbeck2019sampling}.
Since none of these steps depend on specific model properties, model-agnostic interpretation techniques allow us to study the behavior of arbitrary models.
Partial dependence (PD) plots \citep{friedman1991multivariate} and permutation feature importance (PFI) \citep{breiman2001random,fisher2019all} are popular model-agnostic methods for describing the relationship between input features and model outcome on a global level.
PD plots visualize the average effect features have on the prediction, and PFI estimates how much each feature improves the model performance and therefore how relevant a feature is.
However, PD and PFI merely describe the prediction (or classification) function, but lack a theory that connects them to the data generating process.
Treating PD and PFI as statistical estimators (like coefficients in a regression model) would require a theoretical counterpart in the DGP: a ground truth estimand that these interpretation methods are supposed to retrieve.
Furthermore, for proper inference about the DGP, we need to quantify the uncertainty of PD and PFI estimators.
Linear regression models, for example, provide variance estimates for the coefficients, which help to distinguish true effects from randomness and allow confidence interval estimation and hypothesis testing.
Most machine learning approaches, however, do not provide variance estimates for their predictions or model parameters.
Yet, the training process itself can be a relevant source of variance as the trained model heavily depends on the specific training data.


We propose to treat PD and PFI as statistical estimators of a ground truth, which allows us to relate the model interpretation to the data generating process.
In Section~\ref{sec:related-work}, we introduce related work and in Section \ref{sec:background} we introduce notation and background on PD and PFI.
In Section \ref{sec:relating-model-dgp}, we formulate PD and PFI as estimators of (proposed) ground truth estimands in the DGP.
By treating PD and PFI as statistical estimators, we can apply the bias and variance decomposition and identify the different sources of uncertainty.
To reflect the different uncertainty sources, we distinguish between model-PD/PFI and learner-PD/PFI.
The model-PD/PFI (Section \ref{sec:model-pd}) follows the standards definitions of PD and PFI.
We propose confidence intervals and variance estimators for model-PD/PFI and show that they neglect the model variance originating from the training process.
In Section \ref{sec:learner-pd}, we propose the learner-PD and learner-PFI which take the model variance into account, study their statistical biases and propose variance estimators and confidence intervals.
For models that lack variance estimates, multiple model refits are required to capture the variance due to the learning process.
Data size is often a limiting factor, so that model refits are based on resampled data with overlapping observations.
This overlap can lead to an underestimation of variance and thus to confidence intervals that are too narrow.
We leverage a variance correction approach from model performance estimation to improve the variance estimation.
In Section \ref{sec:simulation}, we analyze the coverage of the confidence intervals for learner-PD and learner-PFI with and without the correction.
In the application in Section \ref{sec:application} we demonstrate the use of confidence intervals for PD and PFI and illustrate the importance of taking the model variance into account.

\section{Related Work}
\label{sec:related-work}

For PD plots, model-specific confidence intervals exist that rely on models with inherent variance estimators such as Bayesian additive regression trees \citep{cafri2016understanding,zhao2021causal}.
Furthermore, various applied articles contain computations of PD confidence bands \citep{bair2013multivariable,grange2019using,esselman2015landscape,emrich2016public,page2018evaluation,obringer2018predicting}.
These approaches either quantify only the error due to Monte Carlo approximation or, when they cover model variance, they do not account for underestimation of the variance.
This demonstrates the need for a theoretical underpinning of this inferential tool for practical research.
For PFI and related approaches, multiple suggestions for confidence intervals and variance estimation are available.
Some contributions are specific to the random forest PFI \citep{ishwaran2019standard,archer2008empirical,janitza2018computationally},
for which a test for null importance was proposed by \cite{altmann2010permutation}.

Model-agnostic PFI confidence intervals that are similar to ours are proposed by \cite{watson2019testing,williamson2019nonparametric,williamson2020unified}.
We additionally correct for variance underestimation arising from resampling \citep{nadeau2003inference} and relate the estimators to the proposed ground truth PFI.
An alternative approach for providing bounds on PFI is proposed by \cite{fisher2019all} via Rashomon sets, which are sets of models with similar near-optimal prediction accuracy.
Furthermore, alternative approaches of \enquote{model-free} inference exist \citep{parr2020nonparametric,parr2019stratification,zhang2020floodgate}, which aim to infer properties of the data without an intermediary ML model.

\section{Background and Notation}
\label{sec:background}

We denote the joint distribution induced by the data generating process as $\P_{XY}$, where $X$ is a $p$-dimensional random variable and $Y$ a 1-dimensional random variable.
We describe the true mapping from features $X$ to the target $Y$ with $f(X) = \E[Y|X=x]$.
We denote a single random draw from the DGP with $\xi$ and $\yi$.
A dataset consisting of multiple draws from $\P_{XY}$ will be called $\D_n = \{(x^{(1)},y^{(1)}), \ldots, (x^{(n)},y^{(n)})\}$, where $n$ is the number of samples and with each $(\xi, \yi) \sim \P_{XY}$, $i \in \{1, \ldots,n\}$.
An ML model $\fh$ is a function ($\fh: \Xspace \to \Yspace$) that maps a feature vector to a prediction (e.g. $\Yspace = \R$ for regression).
The model $\fh$ is induced based on a dataset $\D_{n}$, using a loss function  $L: \Yspace \times \R^p \to \R_0^+$.
As the true function $f$ is unknown, the model $\fh$ is interpreted instead of $f$, for example, with PD plots and PFI.
The model $\fh$ is learned by an ML learner $I: \D \times \Lambda \to \Hspace$ that maps from the space of datasets and the space of hyperparameters $\Lambda$ to the function hypothesis space $\Hspace$.
The learning process contains two sources of randomness:
the training data being a random sample from $\P_{XY}$ and (possibly) the inherent randomness of the training process \citep{bouthillier2021accounting}.\footnote{For example, stochastic gradient descent and weight initialization in neural networks or bootstrap and feature sampling in random forests are sources of randomness.}
Thus, a model $\fh$ can be seen as realization of a random variable $\F$ with distribution $\P_{\F}$.
We assume that the model is evaluated with a risk function $\risk(\fh) = \E_{XY} [L(Y, \fh(X))] = \int L(y, \fh(x)) \text{d}\P_{XY}$, based on a loss function $L$.
To get unbiased estimates of the risk, model training and evaluation use  different datasets.
The dataset $\D_n$ is split into $\D_{n_1}$ for model training and $\D_{n_2}$ for evaluation, with $n_1 + n_2 = n$.
The empirical risk is estimated with $\riske(\fh_{\D_{n_2}, \lambda}) := \frac{1}{n_2} \sum_{i=1}^{n_2} L\left(\yi,\fh_{\D_{n_2}, \lambda}(\xi)\right)$.

We distinguish between the "simulation" and the "real world" scenario \citep{hothorn2005design}.
In the simulation scenario, we can generate a quasi-infinite number of datasets, which allows us to refit the model multiple times using fresh data each time.
In the real world setting, we assume that a single dataset of size $n$ is available.
To fit multiple models (of the same class) and to obtain multiple estimates of the risk, resampling techniques such as bootstrapping, cross-validation and repeated subsampling have to be used.
We denote by $B_d$ the set of indices for the training data in the $d$-th split repetition and with $B_{-d}$ the corresponding test data indices, where $B_d \cup B_{-d} = \{1,\ldots,n\}$, $b \in \{1,\ldots,m\}$, and $m$ is the number of models trained with different data.

We distinguish between the interpretation of a single model and the distribution of models produced by a learner.
Often a fixed trained model $\fh$ is the subject of interpretation.
Any interpretation of a fixed model neglects the model variance originating from the learning process.
Often we are interested in extending the interpretation to the distribution of models produced by a learner.
For example, the importance of a feature in a decision tree might be zero because it was never selected for a split.
However,  if we were to train the tree on a slightly different sample from the same distribution, it might obtain a non-zero importance.
A similar distinction between model and learner can be made for performance estimation, where model performance is estimated with a test set, but learner performance requires averaging performance over $m$ repetitions and thus model refits.

\subsection{Partial Dependence (PD)}
\label{sec:pdp}

The partial dependence function \citep{friedman1991multivariate} of a model $\fh$ describes the expected effect of a feature after marginalizing out the effects of all other features.
Partial dependence of a feature set $X_S$, $S\subseteq \{1,\ldots,p\}$ (usually $|S| = 1$) is defined as:
\begin{align}
PD_S = \E_{X_{C}}[\fh(x, X_{C})],
\label{eq:theoretical-pdp}
\end{align}
\noindent
where $X_C$ are the remaining features so that $S \cup C = \{1,\ldots, p\}$ and $S \cap C = \emptyset$.
The PD is estimated using Monte Carlo integration:
\begin{align}
\widehat{PD}_S(x) = \frac{1}{n_2}\sum_{i=1}^{n_2} \fh(x, x_{C}^{(i)})
\label{eq:estimation-pdp}
\end{align}
For simplicity, we write $PD$ instead of $PD_S$, and $\widehat{PD}$ instead of $\widehat{PD}_S$ when we refer to an arbitrary PD.
The PD plot consists of a line connecting the points $\lbrace(x^{(g)}, \widehat{PD}_S(x^{(g)})\rbrace_{g=1}^G$, with $G$ grid points that are  usually equidistant or quantiles of $\P_{X_S}$.
See Figure~\ref{fig:application} for an example of a PD plot.

\subsection{Permutation Feature Importance (PFI)}

The PFI \citep{breiman2001random,fisher2019all} of a model $\fh$ is defined as the increase in loss $L$ when the feature set $X_S$ (usually just one feature) is permuted:
\begin{align}
PFI_S = \E_{\tilde{X}_S X_C Y} [L(Y, \fh(\tilde{X}_S, X_C))] - \E_{XY} [L(Y, \fh(X))],
\label{eq:theoretical-pfi}
\end{align}
where $\tilde{X}_S$ is a random variable based on the distribution of $X_S$.
There are two versions of PFI, the marginal PFI and the conditional PFI, which have different strategies to replace $X_S$ and also different interpretations.
The marginal PFI can be interpreted as the importance of the feature, ignoring dependencies with other features and also ignoring that the data used may differ greatly from the original joint distribution $\P_X$ (extrapolation).
For the marginal PFI we take the expected value over the distribution $\P_{X_S} \cdot \P_{X_CY}$, which means that $\tilde{X}_S$ follows the marginal distribution of $X_S$ and is independent of $X_C$ and $Y$ ($\tilde{X}_S \indep X_C,Y$).
This means that the marginal PFI breaks the association between the feature(s) $X_S$ and the target $Y$, but also between $X_S$ and all other features $X_C$.
For the conditional PFI (cPFI) \citep{molnar2020model,watson2019testing,hooker2019please,candes2018panning}, the expectation is taken over the distribution $\P_{X_S|X_C} \cdot \P_{X_CY}$, so that $\tilde{X}_S$ follows the conditional distribution of $X_S$ given $X_C$ but is still independent of $Y$.
The interpretation of the conditional PFI of a feature is therefore also conditional on all features that are correlated with the feature of interest.
Conditional PFI may be interpreted as the \textit{additional} importance of a feature \textit{given that we already know the other feature values}.

PFI and cPFI are estimated with Monte Carlo integration:
\begin{align}
\widehat{PFI}_S = \frac{1}{n} \sum_{i=1}^{n_2} \left(\frac{1}{l}\sum_{k=1}^l  L(\yi, \fh(\tilde{x}_S^{(k,i)}, x_C^{(i)})) -  L(\yi, \fh(\xi)) \right) ,
\label{eq:model-pfi}
\end{align}
where $\tilde{x}_S^{(k,i)}$ with $k \in \{1,\ldots,l\}$ is the $k$-th sample of $x_S$ for the $i$-th observation.
For the marginal PFI, $\tilde{x}_{S}^{(k,i)}$ can be a permutation of the original vector $x_S$.
The conditional PFI requires a conditional sampling mechanism for the feature, such as subgroups \citep{molnar2020model} or knockoffs \citep{candes2018panning,watson2019testing}.
The estimation of $\widehat{PFI}$ requires unseen data, so that the loss estimates deliver unbiased results \citep{zheng2011cross,Chernozhukov2018}.
If not stated otherwise, mathematical derivations in this paper apply to both marginal and conditional PFI.
We assume that the loss used for PFI can be computed per instance, which excludes losses such as AUC.
See Figure~\ref{fig:application} for a PFI example.
As with PD, we use $PFI$ instead of $PFI_S$ and $\widehat{PFI}$ instead of $\widehat{PFI}_S$s.

\section{Relating Model to Data Generating Process}
\label{sec:relating-model-dgp}

The goal of statistical inference is to gain knowledge about the DGP.
Therefore, the modeler aims to establish relationships between properties of the model and the DGP.
For example, under certain assumptions, the coefficients of a generalized linear model (= model properties) can be related to parameters of the respective conditional distribution defined by the DGP, such as conditional mean and covariance structure (= DGP properties).
Machine learning models such as random forests or neural networks lack such a mapping between learned model parameters and properties of the data generating process.
This lack of counterparts in the DGP make it difficult to interpret complex machine learning models and to draw conclusions about the real world.
Interpretation methods such as PD and PFI provide \textbf{external descriptors} of how features affect the model predictions.
Howerver, PD and PFI are estimators that lack a counterpart estimand in the DGP.
We propose an inference approach for these external descriptors.
We define a ground truth version of PD and PFI directly on the DGP, namely the DGP-PD and the DGP-PFI.
The DGP-PD and the DGP-PFI are defined as the PD and PFI, but applied to the true function $f$ instead of $\fh$.
This means that the DGP-PD becomes the feature effect of features $X_S$ on the underlying function $f$:

\begin{definition}[DGP-PD]
\label{def:dgp-pdp}
The DGP-PD is the PD applied to function $f:\Xspace \mapsto \Yspace$ of the data generating process. $$\text{DGP-PD}(x) = \E_{X_{C}}[f(x, X_{C})]$$
\end{definition}

Similarly, for the DGP-PFI we replace $\fh$ for $f$ and compute the expected losses.
We compute the difference between the loss for the permuted distribution and the loss on the joint distribution.
Since we work with the true $f$, the \enquote{original} loss is the aleatoric uncertainty (without any bias or variance).

\begin{definition}[DGP-PFI]
\label{def:dgp-pfi}
The DGP-PFI is the PFI applied to function $f:\Xspace \mapsto \Yspace$ of the data generating process.  $$\text{DGP-PFI} = \E_{\tilde{X}_S X_C Y}[L(Y,f(\tilde{X}_S, X_C))] - \E_{XY}[L(Y,f(X))]$$
\end{definition}

The function $f$ is usually unknown.
If it were known in an application, we would not need machine learning in the first place.
However, Definitions~\ref{def:dgp-pdp} and \ref{def:dgp-pfi} immediately enable at least two useful applications:
It allows scientists to compare the PD/PFI of a model with the PD/PFI of the DGP \textbf{in simulation studies} and research statistical biases.
More importantly, the ground truth definitions of DGP-PD and DGP-PFI allow us to treat PD and PFI as statistical estimators of properties of the data generating process.

This paper studies PD and PFI as statistical estimators of the ground truth DPG-PD and DGP-PFI, including bias and variance decompositions, and confidence interval estimators.
Whether the estimands themselves are desirable in specific data scenarios and model choices is out-of-scope for this work.
Others have done work in limitations of PFI and PD: For example \cite{molnar2020model,hooker2019please,strobl2008conditional} show that interpretation methods produce misleading results under strongly dependent features (e.g. large correlation between features), \cite{zhao2021causal} assess whether PDs can be used to estimate causal effects, and \cite{groemping2020model} studied whether PDs recover the linear relationship of the DGP when the relationship between target and features is linear.
Extrapolation when features are dependent might be one of the biggest issue for PD and PFI.
As one possible remedy, conditional variants of PDP and PFI  \citep{molnar2020model,fisher2019all,watson2019testing,apley2020visualizing} have been proposed.
For PD, the conditional variant is also called M-Plot \citep{apley2020visualizing} and weights predictions according to how likely their respective feature values are for a given PD grid point.
Our proposed variance and confidence interval estimators and other results apply to both the original and conditional variants of PD and PFI, if not stated otherwise.


\section{Bias-Variance Decomposition}

The definition of DGP-PD and DGP-PFI gives us a ground truth to which the PD and PFI of a model can be compared -- at least in theory and simulation.
The error of the estimation (mean squared error between estimator and estimand) can be decomposed into the systematic deviation from the true estimand (statistical bias) and the variance due to model variance.
PD and PFI are both expectations over the -- usually unknown -- joint distribution of the data.
The expectations are therefore usually estimated from data using Monte Carlo integration, which adds another source of variance to the PFI and PD estimates.
Figure~\ref{fig:pdp-pfi-errors} visualizes the chain of errors that stand between the estimand (DGP-PD,DGP-PFI) and the estimates ($\widehat{PD}$, $\widehat{PFI}$).

\makeatletter
\tikzset{join/.code=\tikzset{after node path={%
\ifx\tikzchainprevious\pgfutil@empty\else(\tikzchainprevious)%
edge[every join]#1(\tikzchaincurrent)\fi}}}
\makeatother
\tikzset{>=stealth',every on chain/.append style={join},
         every join/.style={->}}
\tikzstyle{labeled}=[execute at begin node=$\scriptstyle,
   execute at end node=$]

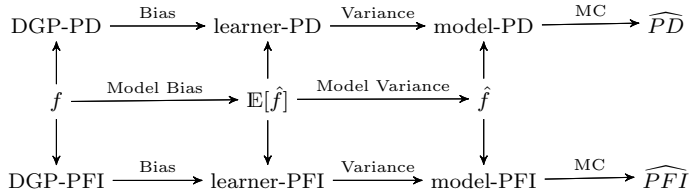
\begin{figure}
\centering
\begin{tikzpicture}
\matrix (m) [matrix of nodes, row sep=1.7em, column sep=4em]
{ DGP-PD & learner-PD & model-PD  & $\widehat{PD}$\\
  $f$ & $\E[\fh]$ & $\fh$ & \\
  DGP-PFI & learner-PFI & model-PFI  & $\widehat{PFI}$\\};
  {[start chain] \chainin (m-1-1);
  \chainin (m-1-2) [join={node[above,labeled] {\text{Bias}}}];
  \chainin (m-1-3) [join={node[above,labeled] {\text{Variance}}}];
  \chainin (m-1-4) [join={node[above,labeled] {\text{MC}}}];
  }
  {[start chain] \chainin (m-2-1);
   { [start branch=A] \chainin (m-1-1)
        [join={node[right,labeled] {}}];}
   { [start branch=B] \chainin (m-3-1)
      [join={node[right,labeled] {}}];}
  \chainin (m-2-2) [join={node[above,labeled] {\text{Model Bias}}}];
     { [start branch=A] \chainin (m-1-2)
        [join={node[right,labeled] {}}];}
   { [start branch=B] \chainin (m-3-2)
      [join={node[right,labeled] {}}];}
  \chainin (m-2-3) [join={node[above,labeled] {\text{Model Variance}}}];
     { [start branch=A] \chainin (m-1-3)
        [join={node[right,labeled] {}}];}
   { [start branch=B] \chainin (m-3-3)
      [join={node[right,labeled] {}}];}
  }

   {[start chain] \chainin (m-3-1);
  \chainin (m-3-2) [join={node[above,labeled] {\text{Bias}}}];
  \chainin (m-3-3) [join={node[above,labeled] {\text{Variance}}}];
  \chainin (m-3-4) [join={node[above,labeled] {\text{MC}}}];
  }
\end{tikzpicture}
\caption{A model $\fh$ deviates from $f$ due to model bias and variance. Similarly $\widehat{PD}$ and $\widehat{PFI}$ estimates deviate from their ground truth versions DGP-PD and DGP-PFI due to bias, variance, and Monte Carlo integration (MC).}
\label{fig:pdp-pfi-errors}
\end{figure}

For the PD, we compare the MSE between the true DGP-PD ($PD_f$ as defined in Equation~\ref{eq:theoretical-pdp}) with the theoretical PD of a model instance $\fh$ ($PD_{\fh}$) at position x.
  \begin{equation*}
    \EF[(PD_{f}(x) - PD_{\fh}(x))^2] = \underbrace{(PD_{f}(x) - \EF[{PD}_{\fh}(x)])^2}_{Bias^2}+ \underbrace{\VF[PD_{\fh}(x)]}_{Variance}
  \end{equation*}
Here, $\F$ is the distribution of the trained models, which can be treated as a random variable.
The bias-variance decomposition of the MSE of estimators is a well known result \citep{geman1992neural}.
For completeness, we provide a proof in Appendix~\ref{sec:proof-bias-variance-pdp}.
Figure~\ref{fig:pdp-uncertainty} visualizes bias and variance of a PD curve, and the variance due to Monte Carlo integration.

Similarly, the MSE of the theoretical PFI of a model (Equation~\ref{eq:theoretical-pfi}) can be decomposed into squared bias and variance.
The proof can be found in Appendix~\ref{sec:proof-bias-variance-pfi}.
\begin{align*}
  \EF[(PFI_{\fh} - PFI_{f})^2] = Bias_{\F}^2[PFI_{\fh}] + \VF[PFI_{\fh}]
\end{align*}

The model variance of PD/PFI stems from variance in the model fit, which depends on the training sample $\D$ and on randomness in the model training such as weight initialization or feature and observation sampling.
When constructing confidence intervals, we have to take into account the variance of PFI and PDP across model fits, and not just the error due to Monte Carlo integration.
As we show in an application (Section~\ref{sec:application}),  whether PD and PFI are based on a single model or are averaged across model refits can impact the interpretation, and especially the certainty of the interpretation.
We therefore distinguish between model-PD/PFI and learner-PD/PFI, which are averaged over refitted models.
Variance estimators for model-PD/PFI only account for variance due to Monte Carlo integration.

\begin{figure}
  \includegraphics[width=\textwidth]{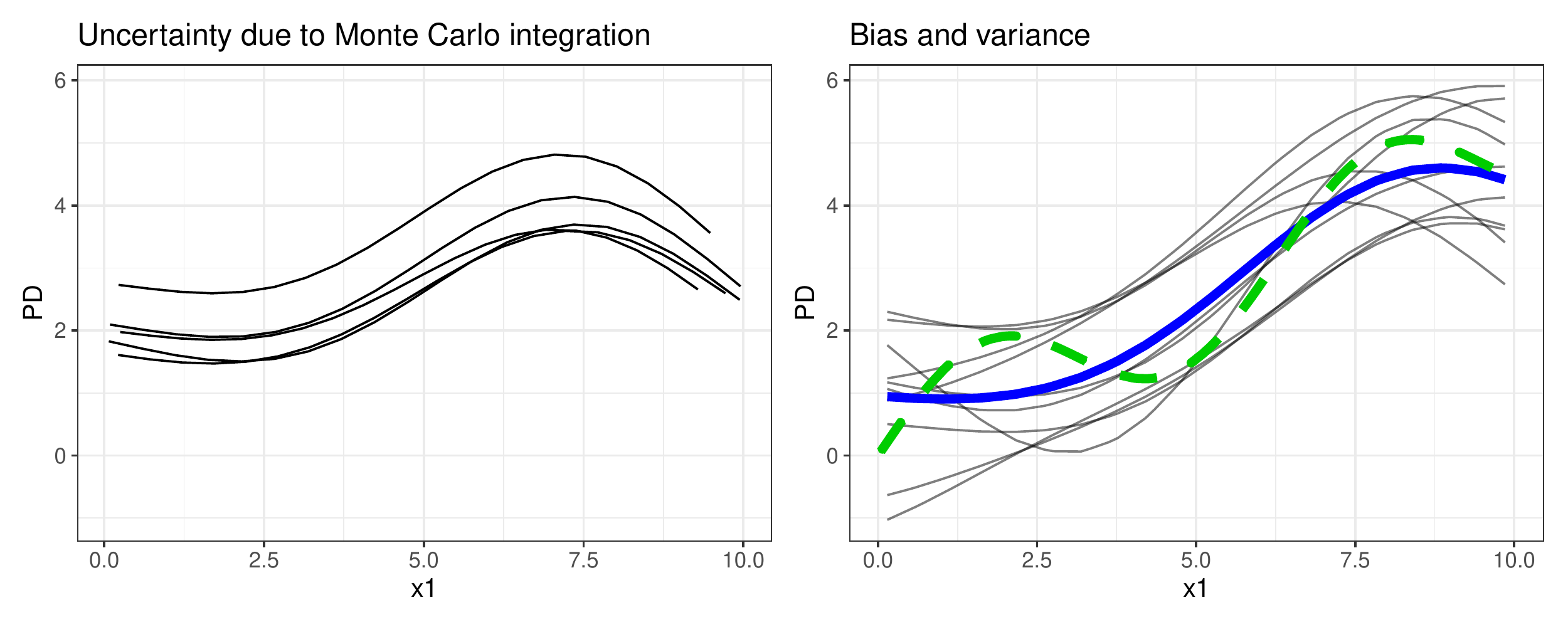}
  \caption{Illustration of bias, variance and Monte Carlo approximation for the PD. Left: Various PDPs using different data for the Monte Carlo integration, but keeping the model fixed. Right: The green dashed line shows the DGP-PD plot of a toy example. Each thin line is the PD plot for the model fitted with a different sample, and the thick blue line is the average thereof. Deviation of the expected PDP from the DGP-PDP are due to bias, deviations of the individual model-PDPs to the expected PDP are due to model variance.}
  \label{fig:pdp-uncertainty}
\end{figure}


\section{Model-PD and Model-PFI}
\label{sec:model-pd}

In this section, we study the model-PD and the model-PFI, and provide variance and confidence interval estimators.
With model-PD and model-PFI, we refer to the original proposals for PD \citep{friedman1991multivariate} and PFI \citep{breiman2001random,fisher2019all} for fixed models.
Conditioning on a given model $\fh$ ignores the model variance due to the learning process.
Only the variance due to Monte Carlo integration can considered in this case.

The model-PD estimator (Equation~\eqref{eq:estimation-pdp}) is unbiased regarding the theoretical model-PD (Equation~\eqref{eq:theoretical-pdp}).
Also, the estimated model-PFI (Equation~\ref{eq:model-pfi}) is unbiased with respect to the theoretical model-PFI (Equation~\ref{eq:theoretical-pfi}).
These findings are general properties of Monte Carlo integration, which state that Monte Carlo integration converges to the integral due to the law of large numbers.
Proofs can be found in Appendix~\ref{sec:model-pdp-unbiased} and \ref{sec:distribution-model-pfi}.
In addition, model-PD  and model-PFI are unbiased estimator of the DGP-PD (Theorem~\ref{th:pdp-unbiased}) and DGP-PFI (Theorem~\ref{th:pfi-unbiased}), under certain conditions.

To quantify the variance due to Monte Carlo integration and to construct confidence intervals, we calculate the variance across the test data instances.
For the model-PD, the variance can be estimated with:
\begin{equation*}
    \widehat{\var}(\widehat{PD}(x)) = \frac{1}{n_2(n_2-1)}\sum_{i=1}^{n_2} \left(\fh(x,\xi_C) - \widehat{PD}(x)\right)^2.
\end{equation*}

Similarly, for the model-PFI the variance is:

\begin{equation*}
  \label{eq:var-model-pfi-term1}
  \widehat{\var}(\widehat{PFI}) = \frac{1}{n_2(n_2 - 1)}\sumin\left(L^{(i)} - \widehat{PFI} \right)^2,
\end{equation*}
where $L^{(i)} = \frac{1}{l} \sum_{k=1}^l L(\yi, \fh(\tilde{x}_S^{(k,i)},\xi_C)) - L(\yi, \fh(\xi))$.

Model-PD and model-PFI are mean estimates of independent samples with estimated variance.
As such, they follow a t-distribution with $n_2 - 1$ degrees of freedom.
This allows us to construct point-wise confidence bands for the model-PD and confidence intervals for the model-PFI, that capture the Monte Carlo approximation uncertainty.
We define point-wise $\alpha$-confidence bands around the estimated model-PD:
\begin{align}
  CI_{\widehat{PD}(x)} = \left[\widehat{PD}(x) - \talpha \sqrt{\widehat{\var}(\widehat{PD}(x))};\widehat{PD}(x) + \talpha \sqrt{\widehat{\var}(\widehat{PD}(x))}\right].
\end{align}
where $\talpha$ is the $1-\alpha/2$ quantile of the t-distribution with $n_2 - 1$ degrees of freedom.
We proceed in the same manner for PFI:

\begin{align}
  CI_{\widehat{PFI}} = \left[\widehat{PFI} - \talpha \sqrt{\widehat{\var}(\widehat{PFI})};\widehat{PFI} + \talpha \sqrt{\widehat{\var}(\widehat{PFI})}\right].
\end{align}

Confidence intervals for model-PD and model-PFI ignore the model variance.
The interpretation, therefore, is limited to variance regarding the Monte Carlo approximation, and we cannot generalize results to the data generating process.
Model-PD/PFI and its confidence bands/intervals are applicable when the focus is a fixed model (e.g. in a model audit).

\section{Learner-PD and Learner-PFI}
\label{sec:learner-pd}

To account for the model variance, we propose the learner-PD and the learner-PFI, which average the PD/PFI over $m$ model fits $\fh_d, d \in \{1,\ldots,m\}$ produced by the same learning algorithm, but trained on different data samples.
The learner-variants are averages of the model-variants, where for each model-PD/PFI the model is repeatedly \enquote{sampled} from the distribution of models.

The learner-PD is therefore the expected PD over the distribution of models generated by the learning process: $\E_F[PD(x)]$.
We estimate the learner-PD with:
\begin{align}
\elPD(x) = \frac{1}{m}\sum_{d=1}^m \frac{1}{|B_{-d}|} \sum_{i \in B_{-d}} \fh_d\left(x, \xi_{C}\right),
\label{eq:pdp-estimator-algorithm}
\end{align}
where $\fh_d$ is trained on sample indices $B_d$ and the PD estimated using samples $B_{-d}$ so that $B_d \cap B_{-d} = \emptyset$.

Following the PD, the learner-PFI is the expected PFI over the distribution of models produced by the learner: $\E_F[PFI]$.
We propose the following estimator for the learner-PFI:
\begin{align}
\elPFI = \frac{1}{m}\sum_{d=1}^m \frac{1}{|B_{-d}|} \sum_{i \in B_{-d}} \left(\Ltmean_{d} - \Li_{d}\right),
\label{eq:pfi-estimator-algorithm}
\end{align}
where losses $\Li_{d} = L(\yi, \fh_d(\xi))$ and $\Ltmean_{d} = \frac{1}{l}\sum_{k=1}^l L(\yi, \fh_d(\tilde{x}^{(k,i)}_S, \xi_C))$ are estimated with data $B_{-d}$ for a model trained on data $B_d$.
Marginal and conditional versions can also be distinguished for the learner-PFI, depending on how $\tilde{X}_S$ was sampled.
A similar estimator has been proposed by \cite{janitza2018computationally} for random forests.

\subsection{Bias of Learner-PD}

The learner-PD is an unbiased estimator of the expected PD over the distribution of models $\F$, since $\E_F[\elPD(x)] = \E_F\left[\frac{1}{m} \sum_{d=1}^{m} \widehat{PD}_d(x)\right] = \frac{m}{m}\E_F[PD_{\fh}(x)] = \E_F[PD_{\fh}(x)]$.
The bias of the learner-PD \textit{regarding the DGP-PD} is linked to the bias of the model.
If the ML model is unbiased, the PDs are unbiased as well.

\begin{theorem}
  Model unbiasedness implies PD unbiasedness: \\
  $\EF[\fh(x)] = f(x) \implies \EF[\mpdp] = \tpdp$
  \label{th:pdp-unbiased}
\end{theorem}
\begin{proof-sketch}
Applying Fubini's Theorem allows us to switch the order of integrals. Further replacing $\EF[\fh]$ with f proves the unbiasedness.
A full proof can be found in Appendix~\ref{sec:proof-pdp-unbiased}.
\end{proof-sketch}

By model bias, we refer to the deviation between the estimated $\fh$ and $f$.
Inductive bias, i.e. the preference of one generalization over another, is necessary for learning \citep{mitchell1980need}.
A wrong choice of inductive bias, such as assuming a linear $\fh$ for a non-linear $f$, leads to deviations of $\fh$ from $f$.
But there are also other reasons why a bias of $\fh$ from $f$ may occur, for example a too small training data size.
We discuss the critical assumption of model unbiasedness further in Section~\ref{sec:discussion}.

\subsection{Bias of Learner-PFI}

The learner-PFI is unbiased regarding the expected learner-PFI over the distribution of models $\F$, since the learner-PFI is a simple mean estimate.
However, unlike the learner-PD, model unbiasedness does not, in general, imply unbiasedness of the learner-PFI \textit{regarding the DGP-PFI}.
In the following, we study the PFI bias when the squared error is used for loss $L$ (L2-loss).

\begin{theorem}
\label{th:pfi-unbiased}
   If model $\fh$ is unbiased with $\E_{\F}[\fh]=f$ and the L2-loss is used, then the conditional model-PFI and conditional learner-PFI are unbiased estimators of the conditional DGP-PFI.
\end{theorem}

\begin{corollary}
  If model $\fh$ is unbiased, the L2-loss is used and the features $X_S$ are independent of features $X_C$, then the marginal model-PFI and marginal learner-PFI are unbiased estimators of the DGP-PFI. If the features are dependent, the following bias is introduced: $PFI_{\fh} - \text{DGP-PFI} = \EXZ[\VF[\fh]] -  \E_X[\VF[\fh]]$.
\end{corollary}

\begin{proof-sketch}
Both $L$ and $\Lt$ can be decomposed into bias, variance, and irreducible error.
Due to the subtraction, the irreducible error vanishes and the differences of biases and variances remain.
Model unbiasedness sets the bias terms to zero, but the difference in variance only becomes zero if either $X_S \indep X_C$ or conditional PFI is used.
The extended proof can be found in Appendix~\ref{sec:proof-bias-variance-l2}.
\end{proof-sketch}

Sampling feature $X_S$ creates a new distribution $(\tilde{X}_{S}, X_C)$, with a (possibly) different variance for a given point across models.
If the variance of $\fh$ changes for $\tilde{X}_S$, this leads to a bias in the PFI estimate.
Besides this bias due to the extrapolation variance, the assumption of model unbiasedness is critical or even unreasonable for regions outside of $\P_{XY}$, since there is no feedback whether the model matches the DGP in these regions.
Furthermore, the DGP might have a probability density of zero for regions of extrapolation.
This means that the marginal PFI for dependent features can have a conceptual problem, as the permutation might create data points that are in conflict with the DGP \citep{hooker2019please,molnar2020model}.\footnote{Imagine a person with a weight of 4kg and a height of 2m.}

Intuitively, the model-PFI and learner-PFI should tend to have a negative bias and therefore underestimate the DGP-PFI.
A model cannot use more information about the target than is encoded in the DGP (except for dependent features in combination with marginal PFI).
However, as Theorem~\ref{th:cpfi-bias} shows the (conditional) PFI can be larger than the DGP-PFI.

\begin{theorem}
\label{th:cpfi-bias}
The difference between the conditional PFI ($cPFI_{\fh}$) and the conditional DGP-PFI ($cPFI_{f}$) of a model $\fh$ is given by:
  \begin{align*}
  cPFI_f - cPFI_{\fh}&=2\E_{X_C}\bigl[\Var_{X_S\mid X_C}[f]-Cov_{X_S\mid X_C}[f,\fh]\bigr].
  \end{align*}
\end{theorem}

\begin{proof-sketch}
For the L2 loss, the expected loss of a model $\fh$ can be decomposed into the expected loss between $\fh$ and $f$ and the expected variance of $Y$ given $X$. Due to the subtraction, the latter term vanishes. The remainder can be simplified using that $Y\indep \tilde{X_S}\mid X_C$ and $P(\tilde{X_S},X_C)=P(X_S,X_C)$.
The extended proof can be found in Appendix~\ref{sec:proof-l2-DGP-model}.
\end{proof-sketch}

However, for an overestimation of the PFI to occur, the expected conditional variance of $\fh$ must be greater than the one of $f$.
Moreover, $\fh$ and $f$ must have a large expected conditional covariance, meaning that $\fh$ has learned something about $f$.

\subsection{Variance Estimation}

The learner-PD and learner-PFI vary due to model variance (refitted models), but also due to using different samples each time for the Monte Carlo integration.
Their variance estimates therefore capture the entire modeling process.
Insofar, learner-PD/PFI along with their variance estimators bring us closer to the DGP-PD/PFI and only the systematic bias remains unknown.

We can estimate this point-wise variance of the learner-PD with:
\begin{equation*}
  \widehat{\var}(\elPD(x)) = \left(\frac{1}{m} + c\right) \cdot \frac{1}{(m - 1)} \sum_{d=1}^m (\widehat{PD}_d(x) - \elPD(x))^2
  \label{eq:mpdpvar}
\end{equation*}
And equivalently for learner-PFI:
\begin{align*}
  \widehat{\var}(\elPFI) &= \left(\frac{1}{m} + c\right)\cdot \frac{1}{(m - 1)} \sum_{d=1}^{m} (\widehat{PFI}_d - \elPFI)^2
\end{align*}
The correction term $c$ depends on the data setting.
In simulation settings that allow us to draw new training and test sets for each model, we can use $c = 0$, yielding the standard variance estimators.
In real world settings, we usually have a fixed dataset of size $n$ and models are refitted using resampling techniques.
Consequently, data are shared by model refits and variance estimators will underestimate the true variance \citep{nadeau2003inference}.
To correct the variance estimate of the generalization error for bootstrapped or subsampled models, \cite{nadeau2003inference} suggested the correction term $c = \frac{n_2}{n_1}$ (where $n_2$ and $n_1$ are sizes of test and training data).
However, the correction remains a rough correction, relying on the strongly simplifying assumption that the correlation between model refits depends only on the number of shared observations in the respective training datasets, and not on the specific observations that they share.
While this assumption is usually wrong, we show in Section~\ref{sec:simulation} that the correction term offers a vast improvement for variance estimation -- compared to using no correction.

\subsection{Confidence Bands and Intervals}
\label{sec:learner-pfi-cis}

Since learner-PD and learner-PFI are means with estimated variance, we can use the t-distribution with $m-1$ degrees of freedom to construct confidence bands/intervals, where $m$ is the number of model fits.
The point-wise confidence band for learner-PD is:

\begin{equation*}
  CI_{\elPD(x)} = \left[\elPD(x) - \talpha \sqrt{\widehat{\var}(\elPD(x))};\elPD(x) + \talpha\sqrt{\widehat{\var}(\elPD(x))}\right],
\end{equation*}

where $\talpha$ is the respective $1-\alpha/2$ quantile of the t-distribution with $m-1$ degrees of freedom.
Equivalently, we propose a confidence interval for the learner-PFI:

\begin{align*}
  CI_{\elPFI} = \left[\elPFI - \talpha \sqrt{\hat{\var}(\elPFI)};\elPFI+ \talpha \sqrt{\hat{\var}(\elPFI)} \right].
\end{align*}

Respecting the model variance can make a difference in the interpretation as we show in the application, Section~\ref{sec:application}.
Resampling strategies make better use of the data, in the sense that a bigger share of the data ends up being used as test data compared to the holdout strategy.

\section{Confidence Interval Coverage Simulation}
\label{sec:simulation}

\newcommand{\nexperiments}{10,000 }
\newcommand{\nsample}{100 }
\newcommand{\ntrain}{67 }
\newcommand{\ntrue}{10,000 }
\newcommand{\nrefits}{15 }
In simulations we compared confidence interval performance between bootstrapping and subsampling, with and without variance correction.
We simulated two data generating processes: a \textit{linear} DGP was defined as $y = f(x) = x_1 - x_2 + \epsilon$ and a \textit{non-linear} DGP as $y = f(x) = x_1 - \sqrt{1 - x_2} + x_3 \cdot x_4 + (x_4/10)^2 + \epsilon$.
All features were uniformly sampled from the unit interval $[0;1]$ and for both DGPs we set $\epsilon \sim N(0,1)$.
We studied the two settings \enquote{simulation} and \enquote{real world}.
In both settings, we trained (each \nrefits times) linear models (lm), regression trees (tree) and random forests (rf), and computed confidence intervals for learner-PD and learner-PFI across the \nrefits refitted models.
In the \enquote{simulation} setting, we sampled  $n \in \{100, 1,000\}$ fresh data points for each model refit, where $63.2\%$ of the data were used for training and the remaining $36.8\%$ for PDP and PFI estimation.

\begin{table}

\caption{\label{tab:sim-ci-coverage-pdp}Coverage Probability of the 95\% PDP Confidence Bands. boot = bootstrap, subs = subsampling, * = with adjustment.}
\centering
\begin{tabular}[t]{llrrrrrr}
\toprule
dgp & model & n & boot & boot* & subs & subs* & ideal\\
\midrule
linear & lm & 100 & 0.41 & 0.89 & 0.34 & 0.82 & 0.95\\
linear & lm & 1000 & 0.41 & 0.89 & 0.33 & 0.80 & 0.95\\
linear & rf & 100 & 0.39 & 0.86 & 0.36 & 0.83 & 0.95\\
linear & rf & 1000 & 0.38 & 0.87 & 0.35 & 0.83 & 0.95\\
linear & tree & 100 & 0.54 & 0.96 & 0.47 & 0.92 & 0.95\\
linear & tree & 1000 & 0.57 & 0.96 & 0.48 & 0.91 & 0.95\\
non-linear & lm & 100 & 0.43 & 0.90 & 0.36 & 0.84 & 0.95\\
non-linear & lm & 1000 & 0.41 & 0.89 & 0.33 & 0.81 & 0.95\\
non-linear & rf & 100 & 0.39 & 0.87 & 0.36 & 0.84 & 0.95\\
non-linear & rf & 1000 & 0.38 & 0.86 & 0.36 & 0.83 & 0.95\\
non-linear & tree & 100 & 0.58 & 0.98 & 0.51 & 0.95 & 0.95\\
non-linear & tree & 1000 & 0.59 & 0.97 & 0.51 & 0.94 & 0.95\\
\bottomrule
\end{tabular}
\end{table}

\begin{table}

\caption{\label{tab:sim-ci-coverage-pfi}Coverage Probability of the 95\% PFI Confidence Intervals. boot = bootstrap, subs = subsampling, * = with adjustment.}
\centering
\begin{tabular}[t]{llrrrrrr}
\toprule
dgp & model & n & boot & boot* & subs & subs* & ideal\\
\midrule
linear & lm & 100 & 0.27 & 0.70 & 0.23 & 0.63 & 0.94\\
linear & lm & 1000 & 0.25 & 0.68 & 0.21 & 0.60 & 0.95\\
linear & rf & 100 & 0.44 & 0.92 & 0.39 & 0.88 & 0.95\\
linear & rf & 1000 & 0.42 & 0.90 & 0.38 & 0.86 & 0.95\\
linear & tree & 100 & 0.52 & 0.97 & 0.42 & 0.90 & 0.95\\
linear & tree & 1000 & 0.42 & 0.90 & 0.34 & 0.81 & 0.95\\
non-linear & lm & 100 & 0.31 & 0.81 & 0.25 & 0.72 & 0.94\\
non-linear & lm & 1000 & 0.25 & 0.67 & 0.21 & 0.59 & 0.95\\
non-linear & rf & 100 & 0.47 & 0.94 & 0.43 & 0.91 & 0.95\\
non-linear & rf & 1000 & 0.41 & 0.89 & 0.38 & 0.86 & 0.95\\
non-linear & tree & 100 & 0.68 & 0.99 & 0.56 & 0.96 & 0.94\\
non-linear & tree & 1000 & 0.58 & 0.97 & 0.46 & 0.92 & 0.95\\
\bottomrule
\end{tabular}
\end{table}

In the \enquote{real world} setting, we sampled $n \in \{100, 1,000\}$ data points \textbf{once} per experiment, and generated $\nrefits$ training data sets using bootstrap (sample size $n$ with replacement, which yields $0.632 \cdot n$ unique data points in expectation) or subsampling (sample size $0.632 \cdot n$ without replacement).
In both settings, learner-PD and learner-PFI plus their respective confidence intervals were computed over the $\nrefits$ retrained models.
We repeated the experiment \nexperiments times and counted how often the estimated confidence intervals covered the expected PD or PFI ($\E_{\fh}[PD]$ and $\E_{\fh}[PFI]$) over the distribution of models $\F$.\footnote{The coverage is not regarding the DGP-PD/PFI, but regarding the expected learner-PD/PFI, as we studied the choices of resampling and correction for the model variance.}
These expected values were computed using \ntrue separate runs.
The coverage estimates were averaged across features per scenario, and, for PD also across grid points ($\{0.1, 0.3, 0.5, 0.7, 0.9\}$ for all features.

Table~\ref{tab:sim-ci-coverage-pfi} and Table~\ref{tab:sim-ci-coverage-pdp} show that in the \enquote{simulation} setting (\enquote{ideal}), we can recover confidence intervals using the standard variance estimation with the desired coverage probability.
However, in the \enquote{real-world}, setting the confidence intervals for both learner-PD and learner-PFI are too narrow across all scenarios and both resampling strategies, when the intervals are based on naive variance estimates.
Some coverage probabilities are especially low, such as for linear models with $30\%-40\%$.

\begin{figure}
  \includegraphics[width=\textwidth]{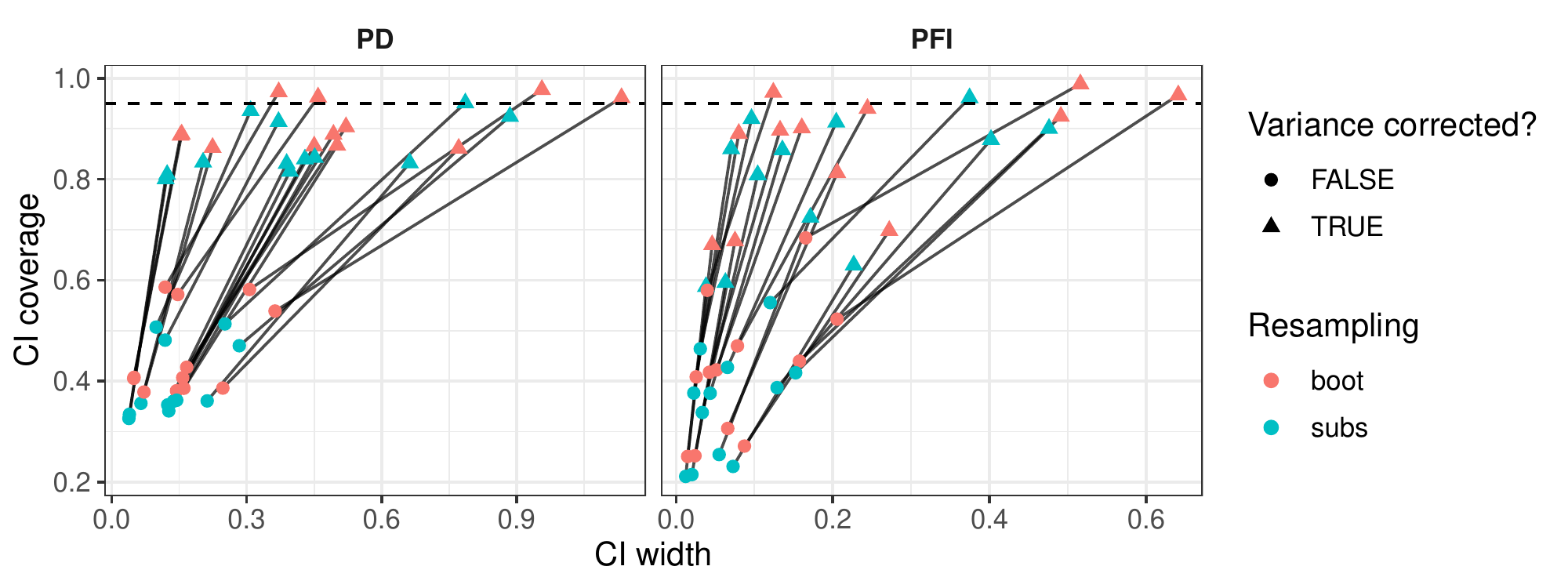}
  \caption{Confidence interval width vs. coverage for \textit{boot}strapping and \textit{subs}ampling, comparing before and after correction. Segments connect identical scenarios.}
  \label{fig:ci-cov-vs-width-adjusted}
\end{figure}

The coverage probabilities drastically improve when the correction term is used, see Figure~\ref{fig:ci-cov-vs-width-adjusted}.
However, in the simulated scenarios, they are still somewhat too narrow.
For the linear model, the confidence intervals were the most narrow with coverage probabilities of around $80\% - 90\%$ for PD and $60\% - 80\%$ for PFI across DGPs and sample sizes.
The PD confidence bands were not much affected by increasing sample size $n$, but the PFI estimates became slightly more narrow in most cases.
In the case of decision trees, the adjusted confidence intervals were sometimes too large, especially for adjusted bootstrap.

\begin{figure}
  \includegraphics[width=\textwidth]{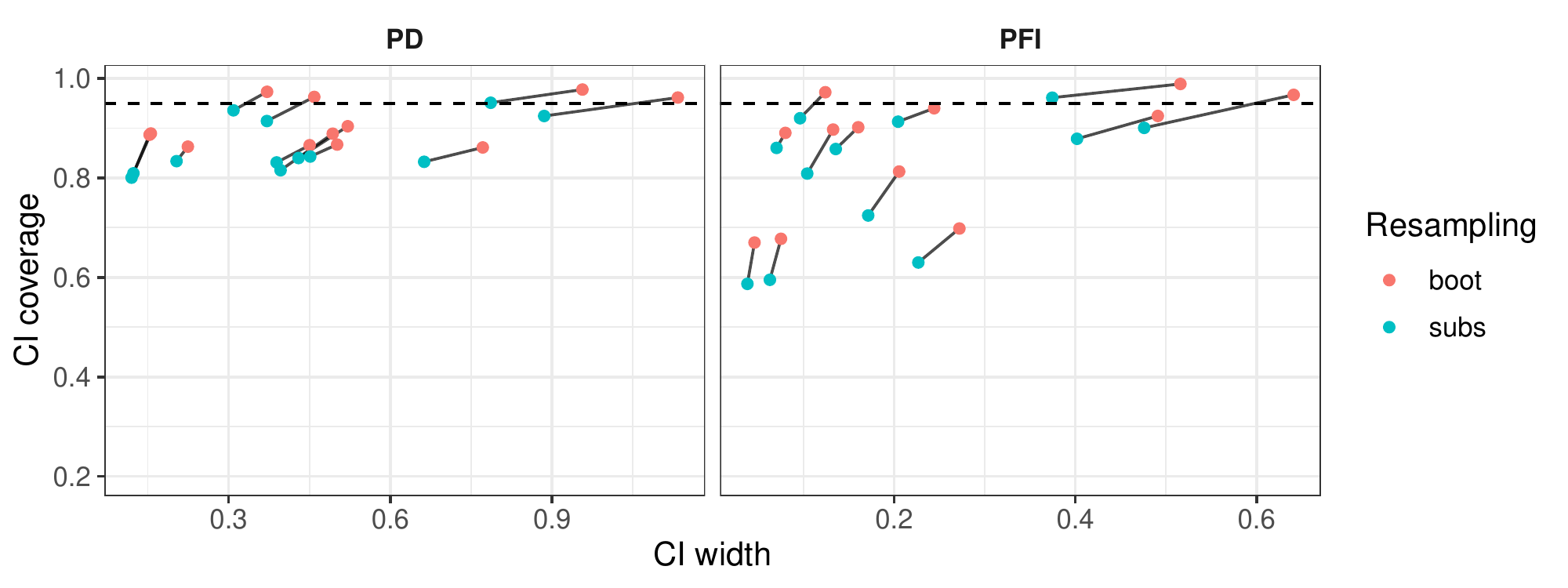}
  \caption{Confidence interval width vs. coverage for \textit{boot}strap and \textit{subs}ampling, both with correction. Segments connect identical scenarios.}
  \label{fig:ci-cov-vs-width}
\end{figure}

\begin{figure}
  \includegraphics[width=\textwidth, trim = {0 0.4cm 0 1cm}, clip]{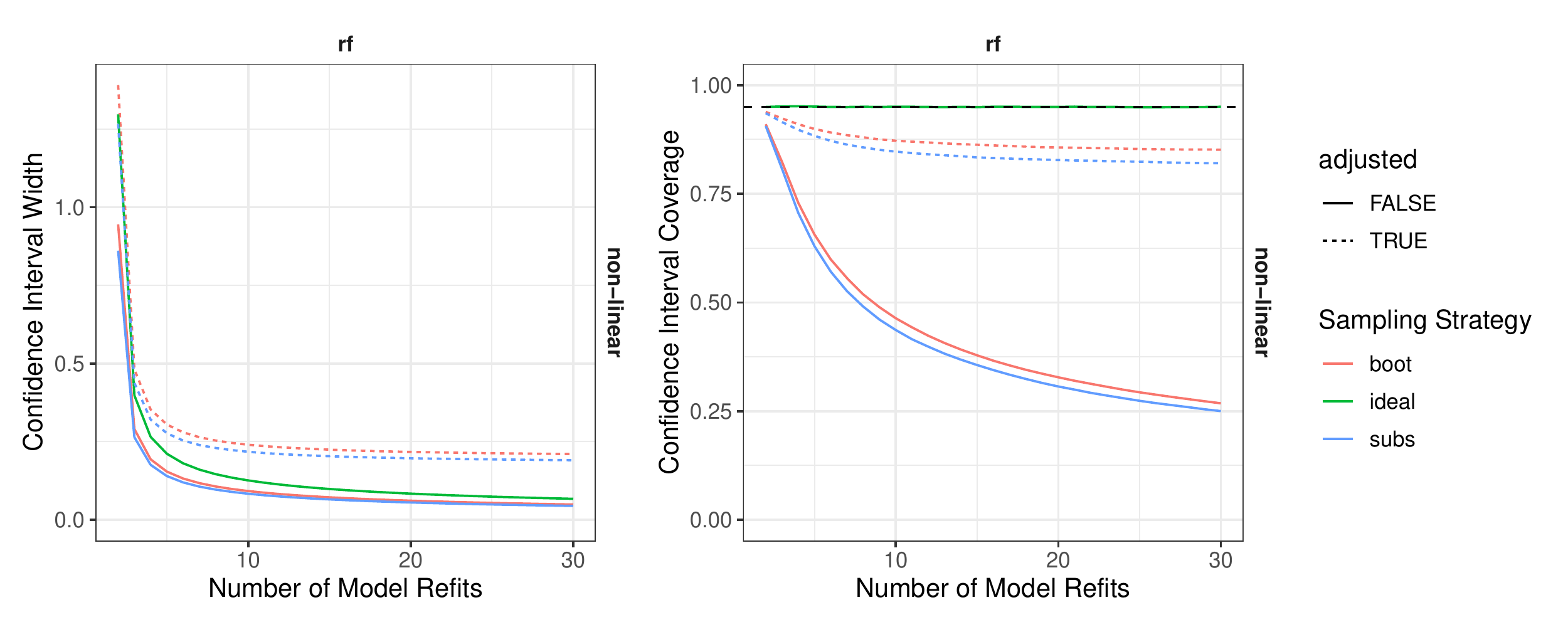}
  \caption{Average PD confidence band width (left) and coverage (right) as a function of number of refitted models for the random forest on the \textit{non-linear} DGP.}
  \label{fig:ci-example}
\end{figure}

Except for trees on the \textit{non-linear} DGP, bootstrap outperformed subsampling in terms of coverage, meaning the coverage was closer to the 95\% level and rather erred on the side of \enquote{caution} with wider confidence intervals (see Figure~\ref{fig:ci-cov-vs-width}).
As recommended in \cite{nadeau2003inference}, we used 15 refits.
We additionally analyzed how the coverage and interval width changed by increasing refits from 2 to 30 and noticed that the coverage worsened with more refits, while the width of the confidence intervals decreased.
Increasing the number of refits comes with an inherent trade-off between interval width and coverage: The more refits are considered, the more accurate the learner-PFI and learner-PD become and also the more certain the variance estimates become, scaling with $1/m$.
But there is a limit to the information in the data, so that additional refits falsely reduce the variance estimate and the confidence intervals become too narrow.
To refit the model 10 - 20 times seemed to be an acceptable trade-off between coverage and interval width, see for example Figure~\ref{fig:ci-example}.
Below $\sim 10$ refits, the confidence intervals were large, and also the mean PD/PFI estimates have a high variance.
Above $\sim 20$ refits, the widths did not decrease by much anymore.
The figures for the other scenarios can be found in Appendix~\ref{app:simulation}.
With our simulation results we could show that confidence intervals using the naive variance estimation (without correction) results in way too narrow intervals.
While the  simple correction term by \cite{nadeau2003inference} does not always provide the desired coverage probability, it is a vast improvement over the naive approach.
We therefore recommend using the correction when computing confidence intervals for learner-PD and learner-PFI -- it is currently the best approach available.
We also recommend refitting the model around 15 times.
For more \enquote{cautious} confidence intervals we recommend using confidence intervals based on resampling with replacement (bootstrap) over sampling wihtout replacement (subsampling).
However, beside wider confidence intervals, the bootstrap requires additional attention when model tuning with internal resampling is used, as data points may otherwise end up in both training and validation data.

\section{Application}
\label{sec:application}

We apply our proposed estimators to predict wine quality \citep{wine} ($n=1599$) from physicochemical features such as alcohol content and acidity.
We compared the performance (mean squared error) of a linear regression model, a regression tree (CART) \citep{cart} and a random forest \citep{breiman2001random} using 15 bootstrap samples (sample size $n$ with replacement).
\begin{figure}
   \centering
  \includegraphics[width=\textwidth,trim = {0 0.4cm 0 0.5cm}, clip]{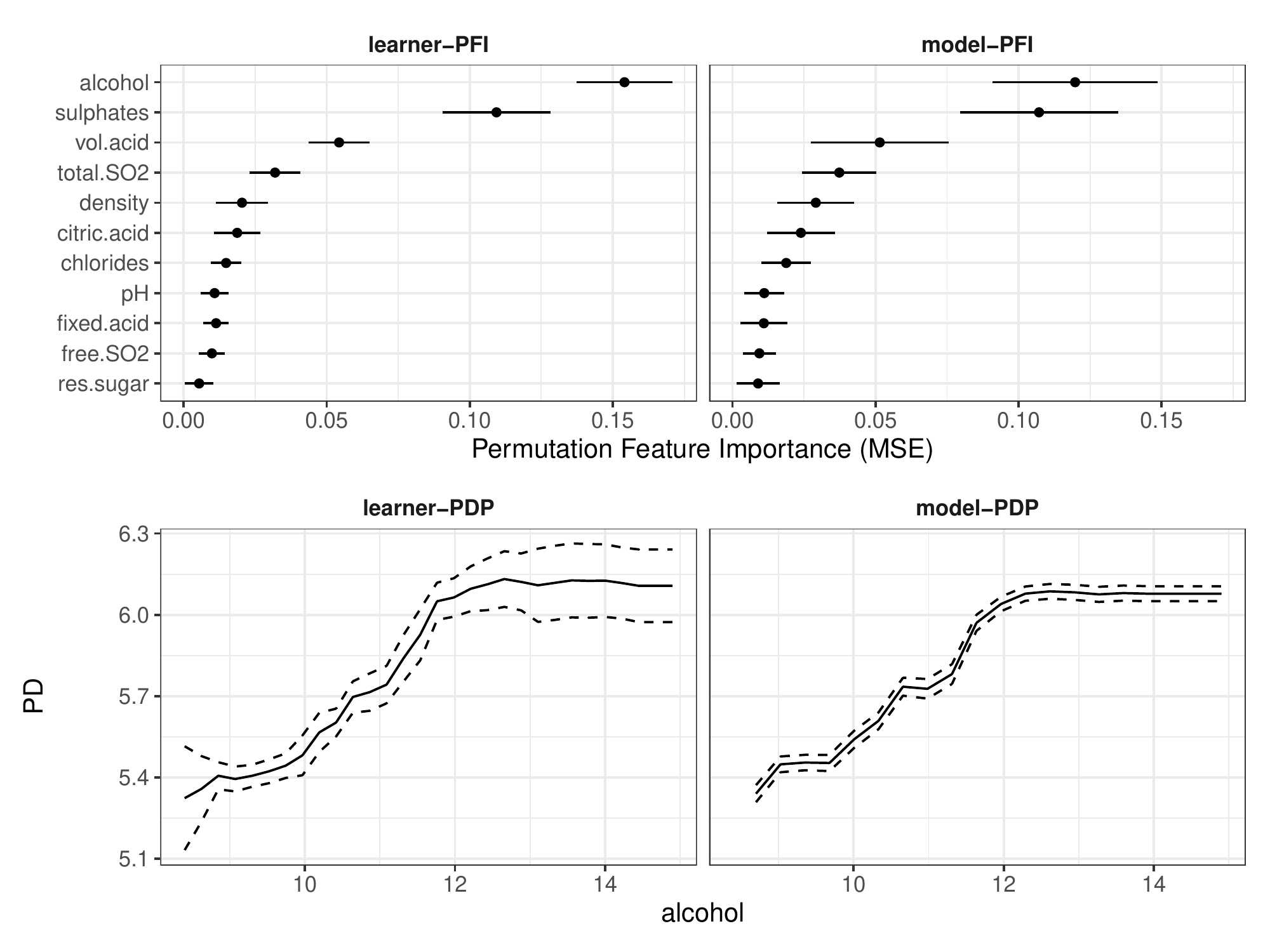}
  \caption{Top: Lerner-PFI and model-PFI with point-wise 95\%-confidence intervals for the random forest. Bottom: Lerner-PDP and model-PDP with point-wise 95\%-confidence bands for the random forest and feature "alcohol".}
  \label{fig:application}
\end{figure}
The MSEs for the different models were: 0.425 (Linear  regression), 0.342 (Random  Forest) and  0.456 (Tree).
The random forest was significantly better than the other models based on an adjusted t-test of the performance difference \citep{nadeau2003inference},
with a 95\% confidence interval of [-0.098;-0.069]
 for the difference to the linear model MSE and [-0.158;-0.071]
 for the difference to the decision tree.
We reused the 15 random forests from the bootstrap to estimate the learner-PD and learner-PFI including their confidence intervals based on adjusted variance estimates.
Figure~\ref{fig:application}, top row, shows that the most important features were alcohol, sulphates and volatile acidity.
The model-PFI quantifies how important each feature was for a fixed random forest, and the confidence intervals show the variance of the approximation of the model-PFI due to Monte Carlo integration.
The model-PFI, however, cannot tell us how much the estimate varies due to model variance.
The learner-PFI quantifies this model variance.
Both model-PFI and learner-PFI gave a similar ordering for the top features.
The learner-PFI shows that alcohol is more important than sulphates (with no overlap in the confidence intervals), for which the model-PFI would suggest that the importance is almost equal.

Figure~\ref{fig:application}, bottom row, shows both the model-PDP and the learner-PDP for the alcohol feature.
Notably, the confidence bands of the learner-PDP are wider than of the model-PDP.
Especially for very low and for high alcohol volumes the models have a high variance.
Neglecting the model variance would mean being overconfident about the partial dependence curve.
In particular, the Monte Carlo approximation error decreases with $1/n$ as the sample size $n$ for PD and PFI estimation increases.
Wrongly interpreted, this can lead to a false sense of confidence in the estimated effects and importance, even though only one model is considered and model variance is ignored.

\section{Discussion}
\label{sec:discussion}

We related the PD and the PFI to the data generating process (DGP), proposed variance and confidence intervals, and discussed conditions for inference.
Our derivations were motivated by taking an external view of the statistical inference process, and postulating that there is a ground truth counterpart to PD/PFI in the data generating process.
To the best of our knowledge, statistical inference via model-agnostic interpretable machine learning is already used in practice, but under-explored in theory.

A critical assumption for inference of effects and importance using interpretable machine learning is unbiasedness of the model.
The model bias is difficult to test, and can be introduced by, e.g. choice of model class, regularization and feature selection.
For example, regularization techniques such as LASSO introduce a small bias \textit{on purpose} \citep{tibshirani1996regression} to decrease model variance and improve predictive performance.
We have to better understand how specific biases affect the prediction function and therefore PD and PFI estimates.
Another crucial limitation for inference of PD and PFI is the underestimation of variance due to data sharing between model refits.
While we could show that a simple correction of the variance \citep{nadeau2003inference} vastly improves the coverage, a proper estimation of the variance remains an open issue.
A promising approach relying on repeated nested cross validation to correctly estimate the variance was recently proposed by \cite{bates2021cross}.
However, this approach is more computationally intensive by up to a factor of 1,000.

\bibliographystyle{spbasic}
\bibliography{Bib}

\newpage

\appendix

\section*{Supplementary Material}

\section{Bias and Variance of PD}
\label{sec:proof-bias-variance-pdp}
The expected squared difference between model-PD and DGP-PD can be decomposed into bias and variance.
\begin{proof}

\begin{align*}
  \EF[(PD_{f} - PD_{\fh})^2] &= \EF[(\tpdp - \mpdp)^2] \\
  &= \EF[\tpdp^2] - 2\EF[\tpdp\cdot\mpdp] \\
  &+ \EF[\mpdp^2] \\
  &= \EF^2[\tpdp] + \underbrace{\VF[\tpdp]}_{=0} \\
  &- 2\tpdp\EF[\mpdp] \\
  &+ \EF^2[\mpdp] + \VF[\mpdp] \\
  &= {\underbrace{(\tpdp- \EF[\mpdp])}_{\text{Bias}}}^2 \\
  &+ \underbrace{\VF[\mpdp]}_{\text{Variance}} \\
  &= {\underbrace{(PD_f - \EF[PD_{\fh}])}_{\text{Bias}}}^2 + \underbrace{\VF[PD_{\fh}]}_{\text{Variance}}
\end{align*}
\end{proof}

\section{Bias and Variance of PFI}
The expected squared difference between model-PFI and DGP-PFI can be decomposed into bias and variance.
\label{sec:proof-bias-variance-pfi}
\begin{proof}
  \begin{align*}
    \EF[(PFI_{\fh} - PFI_{f})^2] &= \EF[PFI_{\fh}^2] + \EF[PFI_{f}^2] \\
     &- 2 \EF[PFI_{\fh} PFI_{f}] \\
     &= \VF[PFI_{\fh}] + \EF[PFI_{\fh}]^2 \\
     & + PFI_{f}^2 - 2 \EF[PFI_{\fh} PFI_{f}] \\
     &= (PFI_{f} - \EF[PFI_{\fh}])^2 + \VF[PFI_{\fh}] \\
     &= Bias^2_{\F}[PFI_{\fh}] + \VF[PFI_{\fh}]
  \end{align*}
\end{proof}

\section{Model-PD Unbiasedness Regarding Theoretical PD}
\label{sec:model-pdp-unbiased}
\begin{proof}
By the law of large numbers, the Monte Carlo integration converges with $n_2 \rightarrow \infty$ to the true integral.
Assuming $n_2$ identically distributed random draws  $X_{C}^{(1)}, \ldots, X_C^{(n_2)} \sim X_{C}$ and model $\fh$, the estimate is:
\begin{align*}
\E_{X_C}[\widehat{PD}(x)] &= \E_{X_{C}}\left[\frac{1}{n_2}\sum_{i=1}^{n_2} \fh(x, X_{C}^{(i)})\right] \\
                          &= \frac{1}{n_2}n_2\E_{X_{C}}[\fh(x, X_{C})] \\
                          &=  PD(x)
\end{align*}
and therefore unbiased for the interval, i.e., the theoretical PD of the model.
\end{proof}

\section{Model-PD Unbiasedness Regarding DGP-PD}
\label{sec:proof-pdp-unbiased}

\begin{proof}
Unbiasedness of the model $\fh$ implies unbiasedness of the model-PD.
\begin{align*}
\EF[\mpdp]  &\overset{Def}{=} \int_{\F} \int_{X_C} \fh(x_S, X_C) d\P(X_C) d\P(\F) \\ 
            &\overset{Fub}{=} \int_{X_C} \int_{\F} \fh(x_S, X_C) d\P(\F) d\P(X_C) \\ 
            &\overset{Def}{=}\E_{X_C}[\EF[\fh]] \\ 
            &\overset{Unbiased}{=}\E_{X_C}[f] 
\end{align*}

Fubini's theorem requires that $\int_{\F,X_C}|\fh|d\P_{\F}\P_{X_C} < \infty$.
This is given when it can be guaranteed that the model predictions have an upper bound $c:|\fh(x)| < c < \infty$.
\end{proof}

\section{Model-PFI Regarding theoretical PFI}
\label{sec:distribution-model-pfi}

\begin{proof}
  As a function of random variables, the loss $L$ itself is a random variable.
  We assume that the loss $L^{(i)}$ of observation $i$ is a sample from the distribution of losses: $L^{(i)} \sim L$ and, similarly for the permuted loss: $\tilde{L}^{(k,i)} \sim \tilde{L}$, where $L^{(i)} = L(\yi, \fh(\xi))$ and $\tilde{L}^{(k,i)} = L(\yi, \fh(\tilde{x}_S^{(k,i)},\xi_C))$.

  The expectation of our estimator is:
  \begin{align*}
    \E_{\tilde{X}_S X_S X_C Y}[\widehat{PFI}_{\fh}] &= \E_{\tilde{X}_S X_S X_C Y}\left[\frac{1}{n_2}\sum_{i=1}^{n_2} (\frac{1}{l}\sum_{k=1}^l (\tilde{L}^{(k,i)} - L^{(i)}))\right] \\
    &= \frac{1}{n_2} n_2 \E_{\tilde{X}_S X_S X_C Y}[((\frac{1}{l} l \tilde{L})  -  L)] \\
    &= \E_{\tilde{X}_S X_C Y}[\tilde{L}] - \E_{X_S X_C Y}[L] \\
    &= PFI_{\fh}
  \end{align*}
  In expectation, we retrieve the theoretical PFI of the model.

\end{proof}

\section{PFI Biases for L2}
\label{sec:proof-bias-variance-l2}

We assume that $L$ is the squared loss $L(y, \fh) = {(y - \fh(x))}^2$ and that $\E[Y|X]$ can be described by $f$ with some additive, irreducible, error $\epsilon$ with $\E(\epsilon) = 0$ and $\var(\epsilon) = \sigma^2$.
To further examine the bias for PFI, we apply the Bias-Variance Decomposition also on the loss itself:
In addition, we use that $\EXY[Y]=\E_X[f(X)]$, $\var_{Y}[Y] = \sigma^2$ and $\E[A^2] = \var[A] + \E[A]^2$.
We first derive the bias-variance decomposition of (i) permuted loss and (ii) original loss and derive from that the expected PFI.

For the permuted loss (i):

\begin{align*}
  \EXYZF[\tilde{L}] &= \EXYZF[(Y - \tilde{\fh})^2]\\
  &= \EXYZ[Y^2 - 2Y\EF[\tilde{\fh}] + \EF[\tilde{\fh}^2]] \\
  &= \EXYZ[Y^2 - 2Y\EF[\tilde{\fh}] + \EF[\tilde{\fh}]^2 + \var_{\F}[\tilde{\fh}]] \\
  &= \var_Y[Y] + \E_{\tilde{X}_SX}[f^2- 2f\EF[\tilde{\fh}] + \EF[\tilde{\tilde{\fh}}]^2 + \var_{\F}[\tilde{\fh}]] \\
  &= \underbrace{\sigma^2}_{\text{Data Var}} + \E_{\tilde{X}_S X}\underbrace{\left[(f - \EF[\tilde{\fh}])^2\right]}_{\text{Bias}^2} + \E_{\tilde{X}_S X}\underbrace{[\var_{\F}[\tilde{\fh}]]}_{\text{Variance}}
\end{align*}

For the original loss (ii):
\begin{align*}
  \EXYF[L] &= \EXYF[(Y - \fh)^2]\\
  &= \EXY[Y^2 - 2Y\EF[\fh] + \EF[\fh^2]] \\
  &= \EXY[Y^2 - 2Y\EF[\fh] + \EF[\fh]^2 + \var_{\F}[\fh]] \\
  &= \var_Y[Y] + \E_X[f^2- 2f\EF[\fh] + \EF[\fh]^2 + \var_{\F}[\fh]] \\
  &= \underbrace{\sigma^2}_{\text{Data Var}} + \E_X\underbrace{\left[(f - \EF[\fh])^2\right]}_{\text{Bias}^2} + \E_{X}\underbrace{[\var_{\F}(\fh )]}_{\text{Variance}}
\end{align*}

The expected PFI for feature $X_S$ then is:
\begin{align*}
    PFI & =  \EXYZF[\tilde{L}] - \EXYF[L]\\
  & \overset{(i)+(ii)}{=}  \sigma^2 + \EXZ\left[(f - \EF[\tilde{\fh}])^2\right] + \EXZ[\var_{\F}(\tilde{\fh})] \\
  & -  (\sigma^2 + \E_X\left[(f - \EF[\fh])^2\right] + \E_X[\var_{\F}(\fh )]) \\
  & = \EXZ\left[(f - \EF[\tilde{\fh}])^2\right] -  \E_X\left[(f - \EF[\fh])^2\right]\\
  & + \EXZ[\var_{\F}[\tilde{\fh}]] -  \E_X[\var_{\F}[\fh]]
\end{align*}

We can derive the same L2 decomposition for the DGP-PFI by replacing $\fh$ with $f$ in the equation above.
This yields $PFI_{f} = \EXZ[(f(X) - f(\tilde{X}_S, X_C))^2]$, since $\VF[f] = \VF[\tilde{f}] = 0$ and $\EF[f]=f$ and $\EF[\tilde{f}]=\tilde{f}$.

The bias of the model-PFI, compared to the DGP-PFI, is:
\begin{align}
    \label{eq:pfi-bias}
PFI_{\fh} - PFI_{f}& = \underbrace{\EXZ[(f - \EF[\tilde{\fh}])^2 - (f - \tilde{f})^2] }_{\text{Permutation Loss Bias}} \\
  & -   \underbrace{\E_X\left[(f - \EF[\fh])^2]\right]}_{\text{(Model Bias)}^2}
  +  \underbrace{\EXZ[\VF[\fh]] -  \E_X[\VF[\fh]]}_{\text{Variance Inflation}}
\end{align}

The permutation loss bias and the squared model bias from the equation above are zero when the model is not biased, i.e., $\fh = f$.
The variance inflation term is zero if $\tilde{X}_S \sim X_S | X_C$, which is the case when conditional PFI is used, or when marginal PFI is used and features $X_S$ are independent from features $X_C$.
If the features in $X_S$ and $X_C$ are dependent, the marginal PFI might be biased, even when the underlying model is unbiased.

\section{DGP-PFI minus model-PFI for L2}
\label{sec:proof-l2-DGP-model}
\begin{align*}
    cPFI_f - cPFI_{\fh} &=\Exyperm[(Y - f)^2] - \Exynorm[(Y- f)^2] \\
    &-\Bigl(\Exyperm[(Y- \fh)^2] - \Exynorm[(Y- \fh)^2]\Bigr)\\
    &=\underbrace{\Bigl(\Exynorm[(Y- \fh)^2]-\Exynorm[(Y- f)^2]\Bigr)}_{\text{T1:=}} \\
    &+\underbrace{\Bigl(\Exyperm[(Y- f))^2]-\Exyperm[(Y- \fh)^2]\Bigr)}_{\text{T2:=}}
\end{align*}
We know that for any $g:X\rightarrow Y$ holds:
\begin{equation*}
    \E_{X,Y}[(Y-g)^2]=\E_X[\Var_{Y\mid X}[Y]] + \E_X[(\E_{Y\mid X}[Y]-g)^2]
\end{equation*}
Since $f=\E_{Y\mid X_S,X_C}[Y]$ we can conclude for our first term T1 that:
\begin{align*}
    \text{T1} &=\Exnorm[\Var_{Y\mid X_S,X_C}[Y]] + \Exnorm[(f-\fh)^2]\\
    &-\Bigl(\Exnorm[\Var_{Y\mid X_S,X_C}[Y]] + \underbrace{\Exnorm[(f-f)^2]}_{=0}\Bigr)\\
    &= \Exnorm[(f-\fh)^2]
\end{align*}
We apply the same trick to T2. Moreover, $Y \indep \tilde{X}_S\mid X_C$.
\begin{align*}
    \text{T2} &=\Experm[\Var_{Y\mid \tilde{X}_S,X_C}[Y]] + \Experm[(\E_{Y\mid \tilde{X}_S,X_C}[Y]-f)^2]\\
    &-\Bigl(\Experm[\Var_{Y\mid \tilde{X}_S,X_C}[Y]] + \Experm[(\E_{Y\mid \tilde{X}_S,X_C}[Y]-\fh)^2]\Bigr)\\
    &=\Experm[(\E_{Y\mid X_C}[Y]-f)^2]-\Experm[(\E_{Y\mid X_C}[Y]-\fh)^2]\\
\end{align*}
If we now set together the two terms again and use in the first step that $P(X_S,X_C)=P(\tilde{X}_S,X_C)$, we get:
\begin{align*}
    \text{T1+T2}&=\E_{X_S X_C}[(f-\fh)^2]+\E_{X_S X_C}[(\E_{Y\mid X_C}[Y]-f)^2]\\
    &-\E_{X_S X_C}[(\E_{Y\mid X_C}[Y]-\fh)^2]\\
    &=\Exnorm\Bigl[f^2-2f\fh+\fh^2+\E_{Y\mid X_C}[Y]^2-2\E_{Y\mid X_C}[Y]f+f^2\\
    &-\E_{Y\mid X_C}[Y]^2+2\E_{Y\mid X_C}[Y]\fh-\fh^2\Bigr]\\
    &=2\Exnorm\Bigl[(f^2-\E_{Y\mid X_C}[Y]f)
    -(f\fh-\E_{Y\mid X_C}[Y]\fh)\Bigr]\\
    &=2\E_{X_C}\Bigl[\E_{X_S\mid X_C}\bigl[(f^2-\E_{Y\mid X_C}[Y]f)-(f\fh-\E_{Y\mid X_C}[Y]\fh)\bigr]\Bigr]\\
    &\overset{*}{=}2\E_{X_C}\Bigl[(\E_{X_S\mid X_C}[f^2]-\E_{Y\mid X_C}[Y]\E_{X_S\mid X_C}[f])\\
    &-(\E_{X_S\mid X_C}[f\fh]-\E_{Y\mid X_C}[Y]\E_{X_S\mid X_C}[\fh])\Bigr]\\
    &\overset{**}{=}2\E_{X_C}\Bigl[(\E_{X_S\mid X_C}[f^2]-\E_{X_S\mid X_C}[f]^2)\\
    &-(\E_{X_S\mid X_C}[f\fh]-\E_{X_S\mid X_C}[\fh]\E_{X_S\mid X_C}[f])\Bigr]\\
    &=2\E_{X_C}\bigl[\Var_{X_S\mid X_C}[f]-Cov_{X_S\mid X_C}[f,\fh]\bigr]
\end{align*}
At *, we use the fact that the random variable $\E_{Y\mid X_C}[Y]$ is measurable by the $\sigma$-Algebra generated from $X_C$ and we are inclined to pull it out of the expectation. In **, we use that from $f=\E_{Y\mid X_S,X_C}[Y]$ follows $\E_{X_S\mid X_C}[f]=\E_{Y\mid X_C}[Y]$.


\section{CI simulation results}
\label{app:simulation}

\newcommand{\pltext}[3]{CI #3 for #1 with n=#2.}

\begin{figure*}
    \centering
    \includegraphics[width=\textwidth]{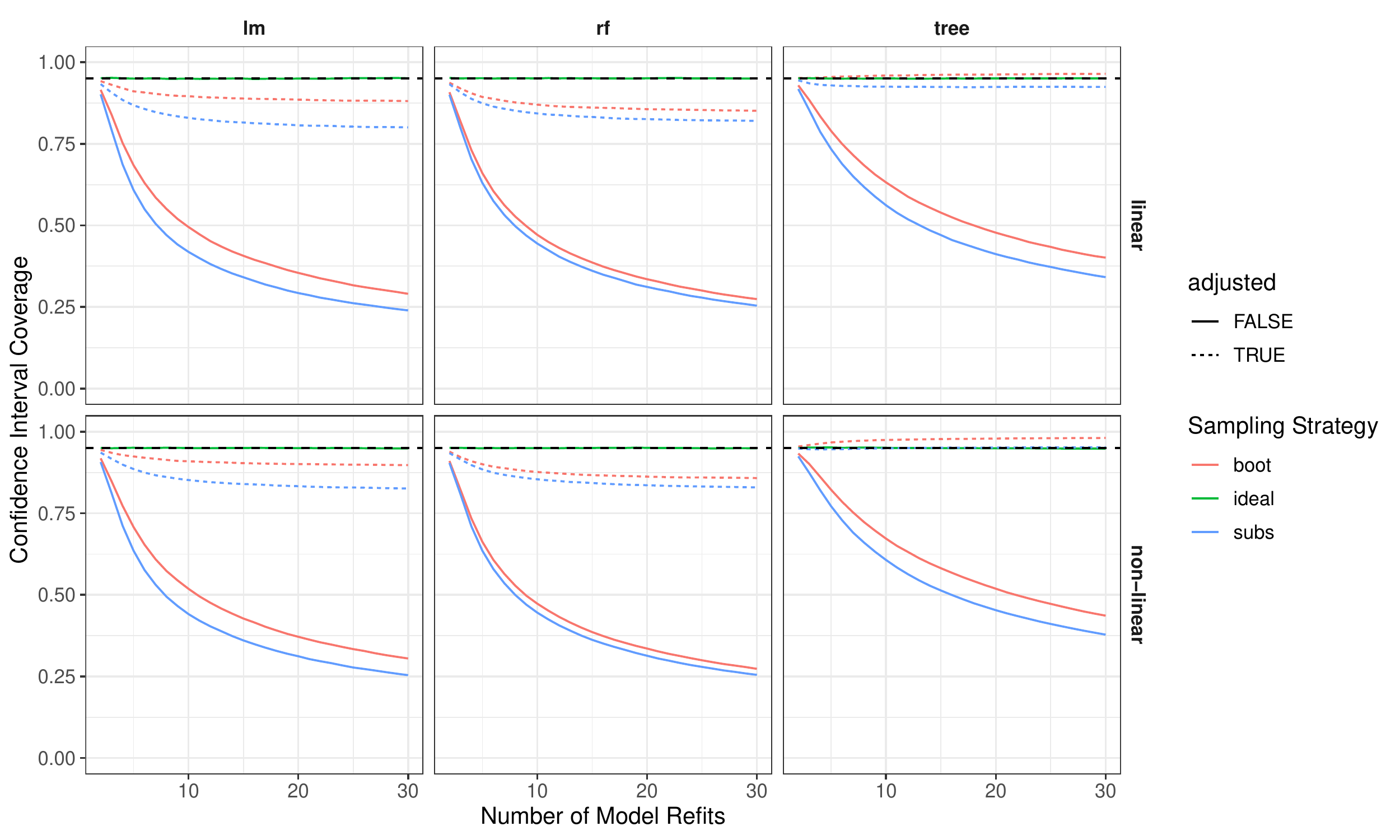}
    \caption{\pltext{PD}{100}{coverage}}
\end{figure*}

\begin{figure*}
    \centering
    \includegraphics[width=\textwidth]{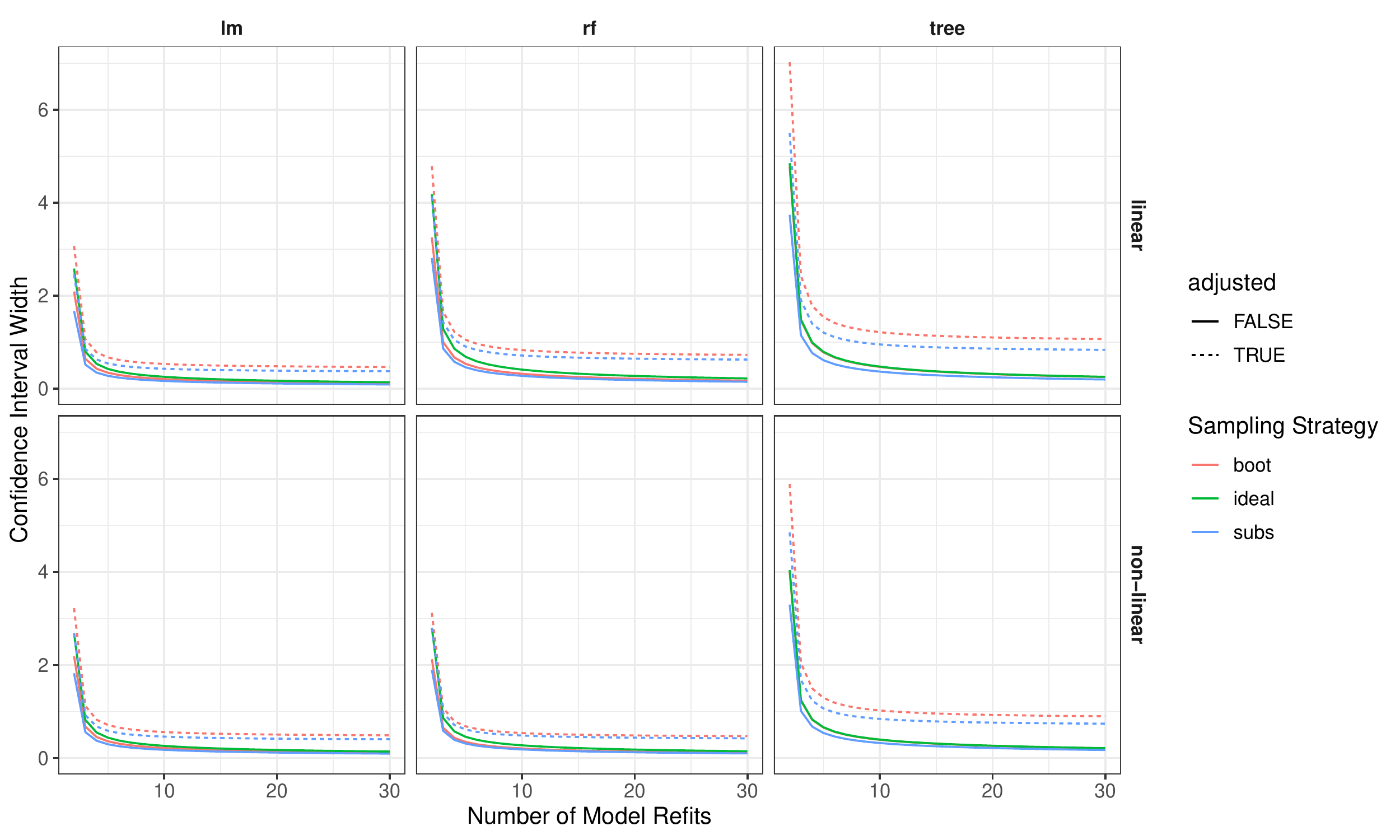}
    \caption{\pltext{PD}{100}{width}}
\end{figure*}

\begin{figure*}
    \centering
    \includegraphics[width=\textwidth]{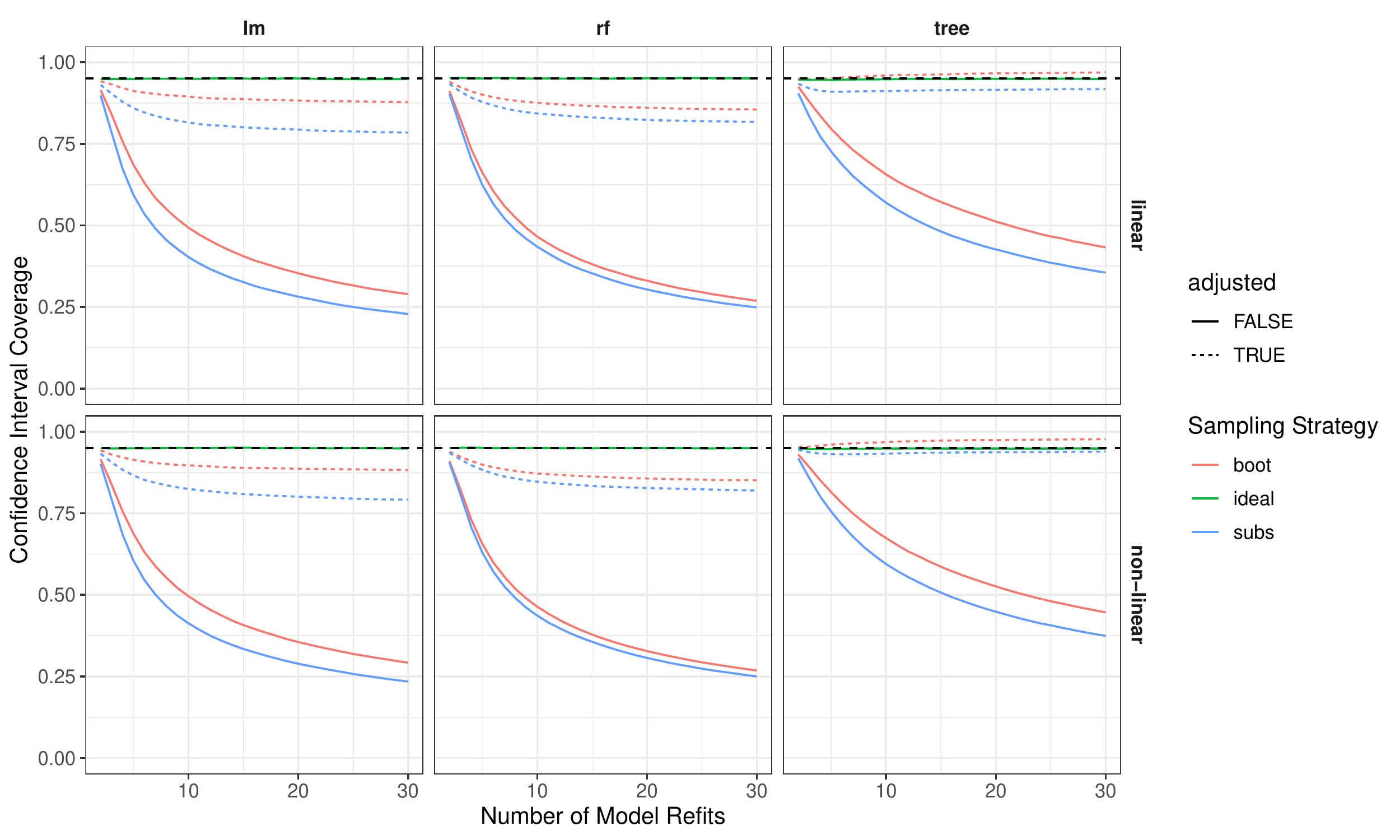}
    \caption{\pltext{PD}{1,000}{coverage}}
\end{figure*}

\begin{figure*}
    \centering
    \includegraphics[width=\textwidth]{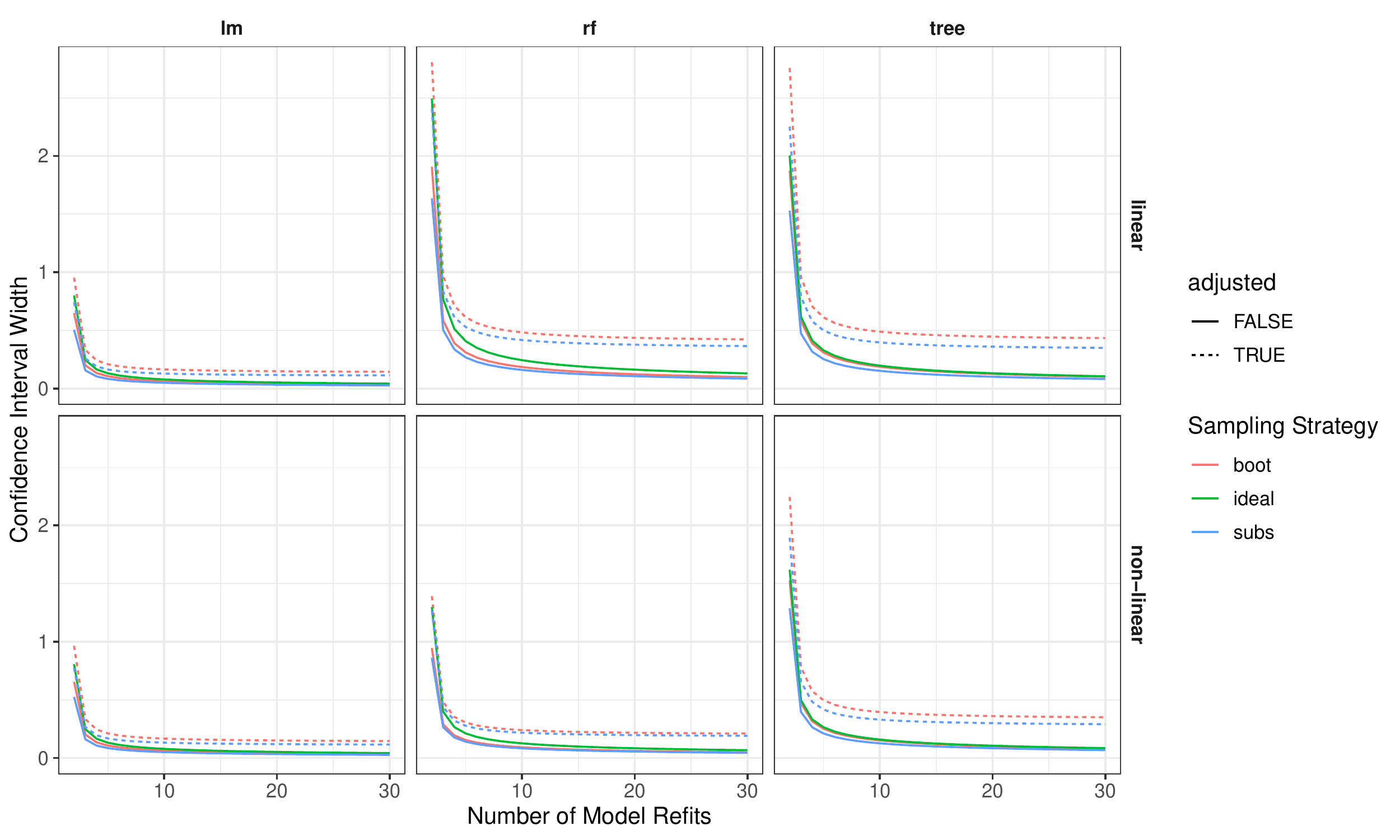}
    \caption{\pltext{PD}{1,000}{width}}
\end{figure*}

\begin{figure*}
    \centering
    \includegraphics[width=\textwidth]{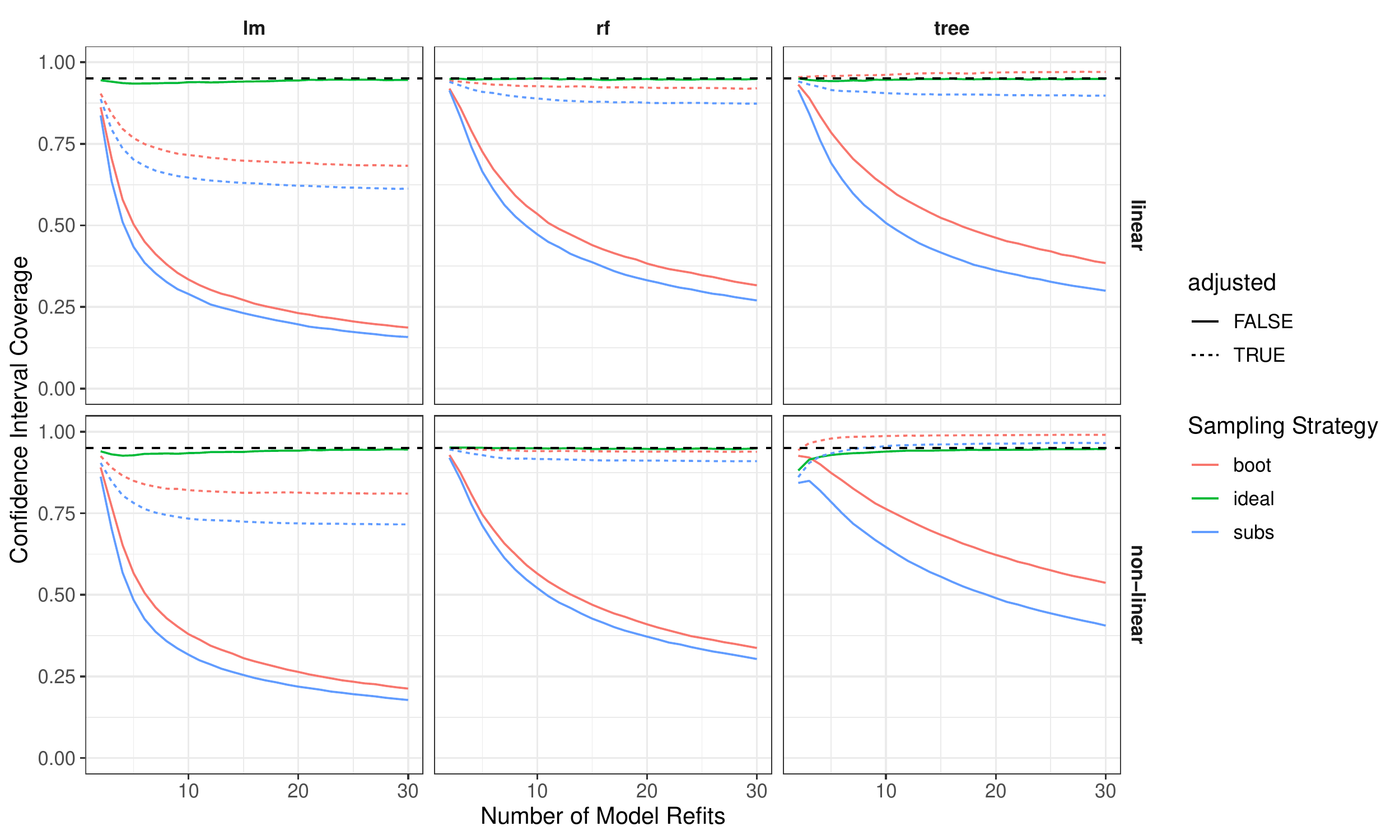}
    \caption{\pltext{PFI}{100}{coverage}}
\end{figure*}

\begin{figure*}
    \centering
    \includegraphics[width=\textwidth]{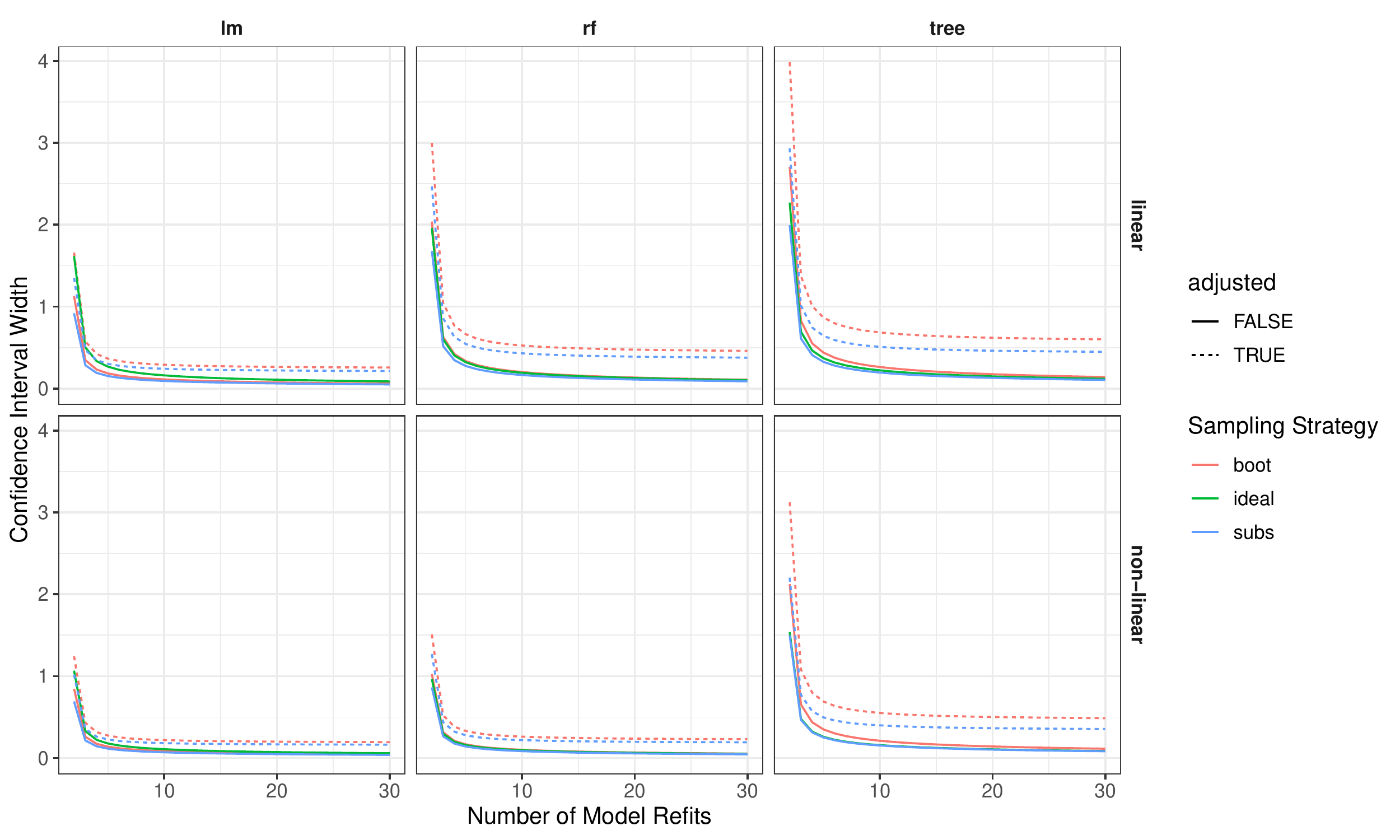}
    \caption{\pltext{PFI}{100}{width}}
\end{figure*}

\begin{figure*}
    \centering
    \includegraphics[width=\textwidth]{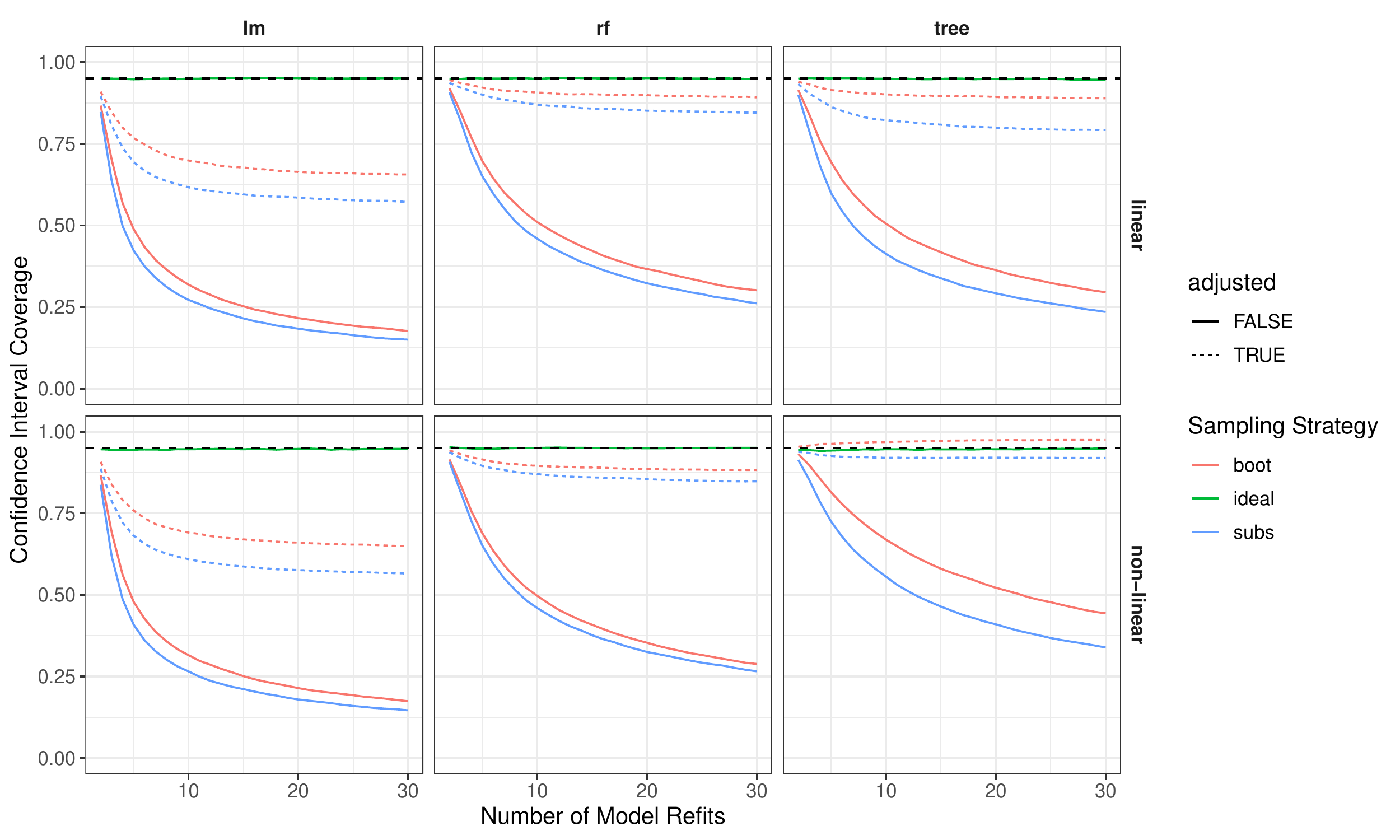}
    \caption{\pltext{PFI}{1,000}{coverage}}
\end{figure*}

\begin{figure*}
    \centering
    \includegraphics[width=\textwidth]{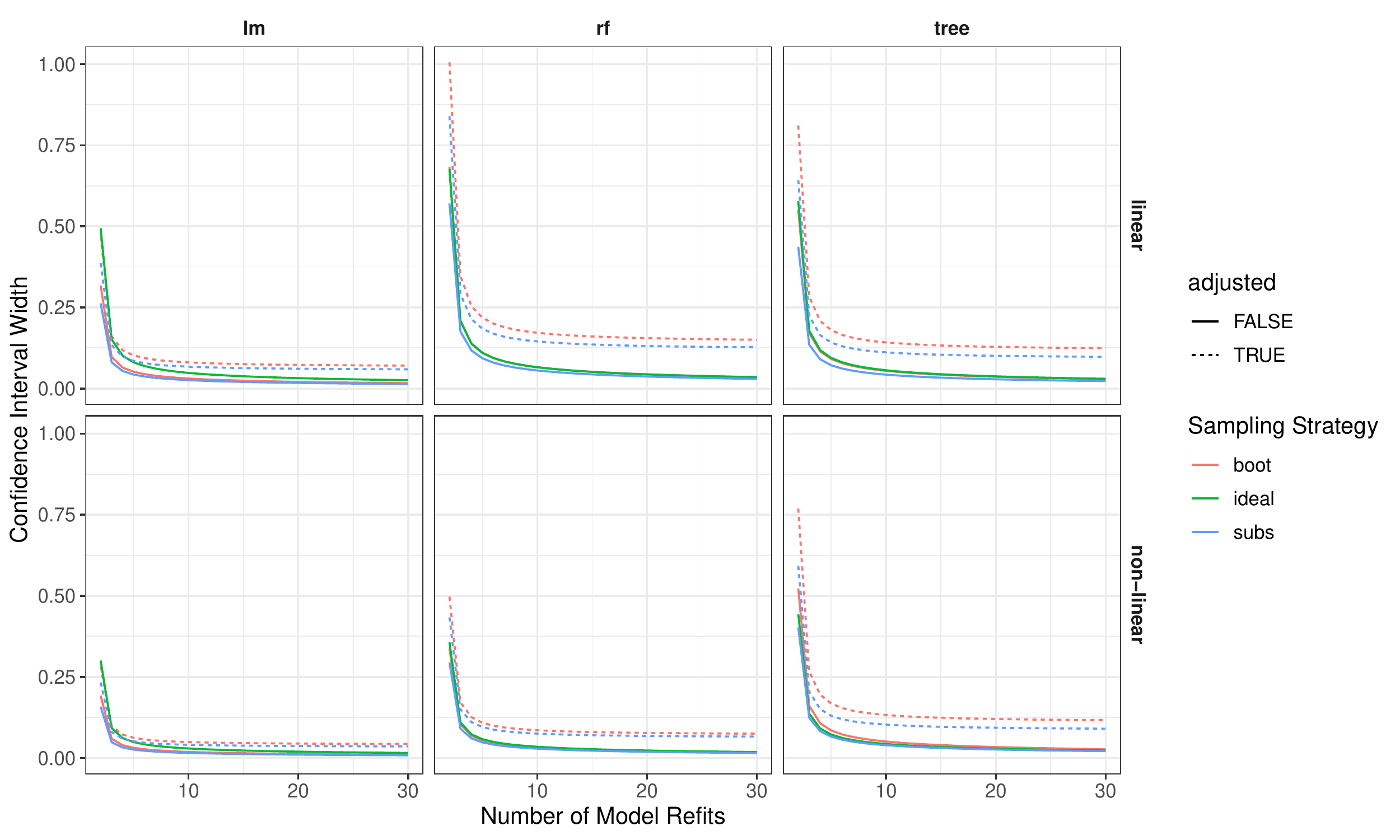}
    \caption{\pltext{PFI}{1,000}{width}}
\end{figure*}

\end{document}